\pdfoutput=1

\documentclass[10pt,twocolumn,letterpaper]{article}

\usepackage{iccv}
\usepackage{times}
\usepackage{amsmath}
\usepackage{amssymb}
\usepackage{caption}
\usepackage{enumitem}
\usepackage{epsfig}
\usepackage{float}
\usepackage{graphicx}
\usepackage{grffile}
\usepackage{listings}      
\usepackage{multirow}
\usepackage{multicol}
\usepackage{ragged2e}
\usepackage{subcaption}
\usepackage{tabularx}
\usepackage{verbatim}
\usepackage{wrapfig}
\usepackage[accsupp]{axessibility} 

\usepackage[pagebackref=true,breaklinks=true,letterpaper=true,colorlinks,bookmarks=false]{hyperref}
\usepackage{cleveref}
\usepackage{lipsum}

\iccvfinalcopy 

\def\iccvPaperID{11164} 

\newcommand{\handle}{\noindent{\texttt{Omnidata} }}

\newcommand\blfootnote[1]{%
  \begingroup
  \renewcommand\thefootnote{}\footnote{#1}%
  \addtocounter{footnote}{-1}%
  \endgroup
}

\ificcvfinal\fi

\begin{document}

\setlist[enumerate]{noitemsep}
\setlist[enumerate,2]{label={\alph*}),leftmargin=0em}

{ 
    \title{
        Omnidata: A Scalable Pipeline for Making\\ Multi-Task Mid-Level Vision Datasets from 3D Scans\\
        \vspace{-5mm}

        }
    


    \author{
        
    \hspace{-3mm}
    Ainaz Eftekhar$^{\dagger *}$\\
    \and
    Alexander Sax$^{\ddagger *}$\\
    \and
    Roman Bachmann$^{\dagger}$\\
    \and
    Jitendra Malik$^{\ddagger}$\\
    \and
    Amir Zamir$^{\dagger}$ \\
    }


    \twocolumn [{%
        \renewcommand\twocolumn[1][]{#1}%
        \maketitle
        \centering
        \vspace{-4mm}
        \hspace{1mm}
        \includegraphics[width=1.00\linewidth]{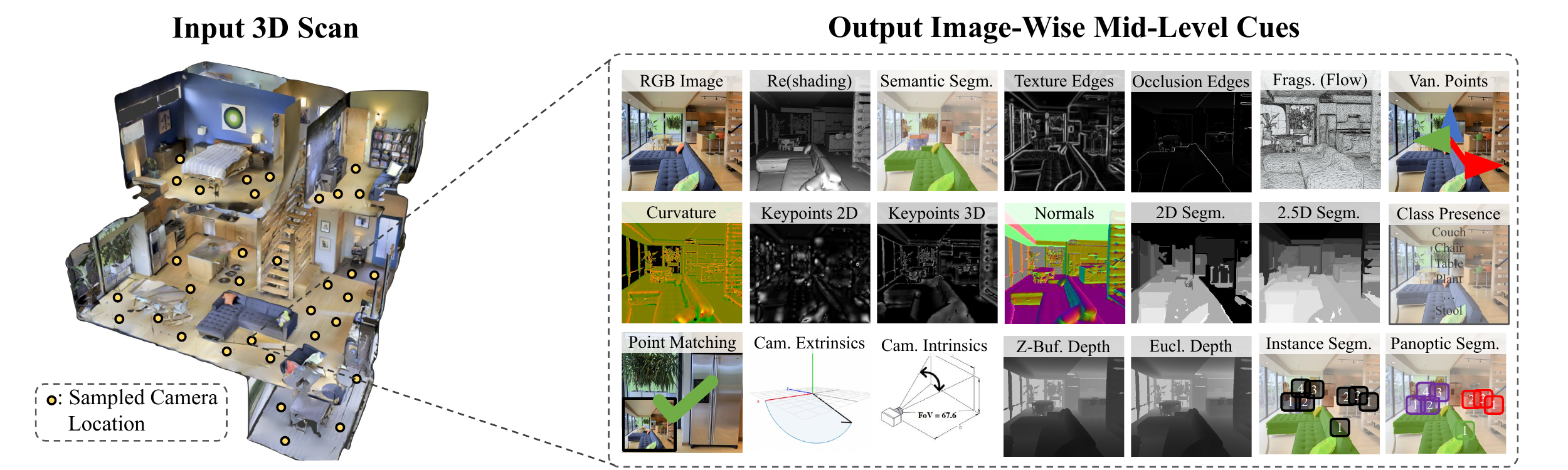}
        \captionof{figure}{\footnotesize{\textbf{Generating `steerable' datasets of mid-level tasks from real-world 3D scans}. \textbf{(Left:)} The proposed pipeline generates dense camera positions and points-of-interest in the input space and \textbf{(right)} renders 21 image-wise mid-level cues by default. Models trained on a starter dataset of this programmatically generated data roughly matches human performance for surface normal estimation on OASIS~\cite{chen2020oasis} (see Sec.~\ref{sec:starter}).
        }}%
        \vspace{8mm}
        \label{fig:pull}
    }]

    \ificcvfinal\thispagestyle{empty}\fi


    \begin{abstract}
    \vspace{-2.0mm}
    This paper introduces a pipeline to parametrically sample and render multi-task vision datasets from comprehensive 3D scans from the real world. Changing the sampling parameters allows one to ``steer'' the generated datasets to emphasize specific information. In addition to enabling interesting lines of research, we show the tooling and generated data suffice to train robust vision models.
    Common architectures trained on a generated starter dataset reached state-of-the-art performance on multiple common vision tasks and benchmarks, despite having seen no benchmark or non-pipeline data. The depth estimation network outperforms MiDaS and the surface normal estimation network is the first to achieve human-level performance for in-the-wild surface normal estimation---at least according to one metric on the OASIS benchmark.
    
    The Dockerized pipeline with CLI, the (mostly python) code, PyTorch dataloaders for the generated data, the generated starter dataset, download scripts and other utilities are available \href{https://omnidata.vision}{through our project website}. 
    \vspace{-3.0mm}
    \end{abstract}

}

\blfootnote{\hspace{-1.5mm}$^{*}$Equal contribution.}

\section{Introduction}
\label{sec:intro}


This paper introduces a pipeline to bridge the gap between comprehensive 3D scans and 
static vision datasets. Specifically, we implement and provide a platform that takes as input one of the following:
\vspace{-0.3em}
\begin{itemize}\setlength\itemsep{-0.5em}
    \item a textured mesh,
    \item a mesh with images from an actual camera/sensor,
    \item a 3D pointcloud and aligned RGB images,
\end{itemize}
\vspace{-0.3em}
and generates a multi-task dataset with as many cameras and images as desired to densely cover the space. For each image, there are 21 different default mid-level cues, shown in Fig.~\ref{fig:pull}. The software makes use of Blender~\cite{blender3d}, a powerful physics-based 3D rendering engine to create the labels, and exposes complete control over the sampling and generation process. With the proliferation of reasonably-priced 3D sensors (e.g. Kinect, Matterport, and the newest iPhone), we anticipate an increase in such 3D-annotated data.

In order to establish the soundness for training computer vision models, we used our pipeline to annotate several existing 3D scans and produce a medium-size starter dataset of mid-level cues. Samples of the data and different cues are shown in Fig.~\ref{fig:starter_dataset_modalities}. Standard models trained on this starter dataset achieve state-of-the-art performance for several standard computer vision tasks. For surface normal estimation, a standard UNet~\cite{Ronneberger15Unet} model trained on this starter dataset yields human-level surface normal estimation performance on the in-the-wild dataset OASIS~\cite{chen2020oasis}, even though the model never saw OASIS data during training. For depth estimation, our DPT-Hybrid~\cite{Ranftl2021} is comparable to or outperforms state-of-the-art models such as MiDaS DPT-Hybrid~\cite{ranftl2019towards, Ranftl2021}. The qualitative performance of these networks (shown in Figs.~\ref{fig:depth-oasis},~\ref{fig:surface_normals}) is often better than the numbers suggest, especially for fine-grained details.

We further provide an ecosystem of tools and documentation around this platform. Our project website contains links to a Docker containing the annotator and all necessary libraries, PyTorch~\cite{NEURIPS2019_9015} dataloaders to efficiently load the generated data, pretrained models, scripts to generate videos in addition to images, and other utilities.

We argue that these results should not be interpreted narrowly. The core idea of the platform is that the ``sectors of the ambient [light-field] array are not to be confused with temporary \emph{samples} of the array" (J. J. Gibson~\cite{Gibson1966}). That is, static images only represent single samples of the entire 360-degree panoramic light-field environment surrounding an agent. How an agent or model samples and represents this environment will affect its performance on downstream tasks. The proposed platform in this paper is designed to reduce the technological barriers for research into the effect of data sampling practices and into the interrelationships between data distribution, data representation, models, and training algorithms. We discuss directions here and analyze a few illustrative examples in the final section of the paper.

First, the pipeline proposed in this paper provides a possible pathway to understand such sampling effects. That is, the rendering pipeline offers complete control over (heretofore) fixed design choices such as camera intrinsics, scene lighting, object-centeredness~\cite{Purushwalkam2020Demystifying}, the level of ``photographer's bias''~\cite{azulay2018deep}, data domain, and so on. This makes it possible to run intervention studies (e.g. A/B tests), without collecting and validating a new dataset or relying on a post-hoc analysis. As a consequence, this provides an avenue for a computer vision ``dataset design guide''.

Second, vision is about much more than semantic recognition, but our datasets are biased towards that as the core problem. The best-studied, most diverse and largest dataset ($>$10M images) generally contains some form of textual/class labels~\cite{imagenet, Sun2017RevisitingUE} and only RGB images. On the other hand, datasets for most non-classification tasks remain tiny by modern standards. For example, the indoor scene dataset NYU~\cite{silberman2012indoor}, still used for training some state-of-the-art depth estimation models~\cite{yin2018geonet}, contains only 795 training images---all taken with a single camera. The pipeline presents a way to generate datasets of comparable quality for non-recognition tasks.

Third, the generated data allows ``matched-pair experimental design" that simplifies study into the \emph{interrelationships} of different tasks, since the pipeline produces labels for every sample. In particular, it helps to avoid issues like the following: suppose a model trained for object classification on ImageNet transfers to COCO~\cite{MSCoco} better than a model trained for depth estimation on NYU~\cite{silberman2012indoor}--is that due to the data domain, the training task, the diversity of camera intrinsics, or something else? 

Existing matched-pair datasets usually focus on a single domain (indoor scenes~\cite{taskonomy2018, silberman2012indoor, 2d3ds,replica19arxiv}, driving~\cite{Dosovitskiy17,cordts2016cityscapes}, block-worlds~\cite{johnson2017clevr}, etc.) and contain few cues~\cite{cordts2016cityscapes,silberman2012indoor,2d3ds,replica19arxiv}. The provided starter dataset may be a better candidate for this research than these existing datasets, since it contains over 14.5 million images from different domains (more than the full ImageNet database), contains many different cues (e.g. for depth, surface normals, curvature, panoptic segmentation, and so on), and models trained on this dataset reach excellent performance for several tasks and existing benchmarks. We demonstrate the value of such matched-pairs data in Sec.~\ref{sec:mtl_exp},

Though our pipeline is designed to facilitate understanding the principles of dataset design, vision beyond recognition, the interrelationships between data, tasks, and models, this paper does not extensively pursue those questions themselves. It provides a few analyses, but these are merely intended as illustrative examples. Instead, the paper introduces tooling designed to facilitate such research as 3D data becomes more widely available and the capture technology improves. \href{http://omnidata.vision}{On our website}, we provide a documented, open-sourced, and Dockerized annotator pipeline with a convenient CLI, runnable examples, a live demo, the starter dataset, pretrained models, PyTorch dataloaders, and code for the paper (including annotator and models).
\begin{figure*}
    \centering
	\includegraphics[width=\linewidth]{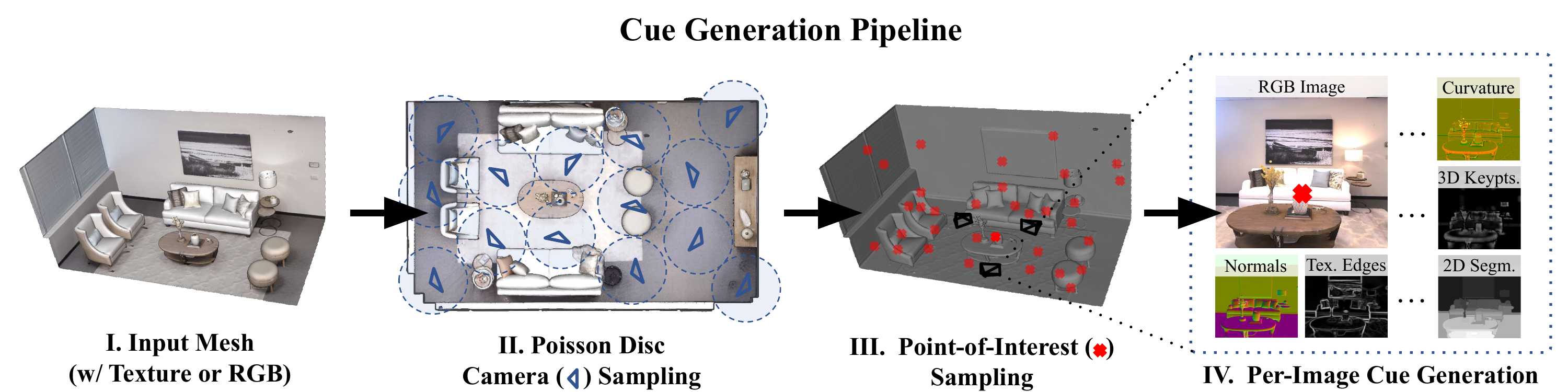}
    \vspace{-0.3cm}		
	\caption{\footnotesize{\textbf{Overview of the generation pipeline.} (\textbf{I}) Given a textured mesh (or other options discussed in Sec.~\ref{sec:generation_pipeline}), our pipeline (\textbf{II}) generates dense camera locations, (\textbf{III}) generates points-of-interest subject to multi-view constraints and (\textbf{IV}) produces 21 different mid-level cues for each (shown in Fig.~\ref{fig:pull}).}\vspace{-5mm}}
	\label{fig:methodology}
\end{figure*}

\section{Related Work}\label{sec:related}

In this section we examine related datasets and other approaches. A thorough review would take more space than we have, so we restrict our attention to only the most relevant groupings.

\textbf{Static 3D Datasets.} 
The past few years have witnessed an uptick in the number of mesh-based datasets, thanks largely to the availability of reasonably-priced 3D scanners. Each dataset in the current crop, though, usually consists of scenes in a restricted domain. Prominent examples of indoor building datasets include Stanford Building Dataset (S3DIS)~\cite{armeni20163d}, Matterport3D~\cite{Matterport3D}, Taskonomy~\cite{taskonomy2018}, Replica~\cite{replica19arxiv}, 2D-3D-Semantic~\cite{armeni2017joint}, Habitat-Matterport~\cite{habitat19iccv}, and Hypersim~\cite{roberts:2020}. Other datasets contain primarily outdoor scenes, usually driving--such as CARLA~\cite{Dosovitskiy17}, GTA5~\cite{Richter_2016_ECCV}--or other narrow domains such as the aptly-named Tanks and Temples~\cite{Knapitsch2017} dataset. Models trained on such scene-level views often do not generalize to object-centric views (see Fig.~\ref{fig:surface_normals}), but existing datasets with high-resolution object meshes do not include 2D images samples~\cite{gsoignition, calli2015benchmarking}. 

Other recent datasets aim to link diverse monocular 2D images and corresponding 3D meshes, but take the reverse approach of this paper by using hand-annotation to create meshes from single-view in-the-wild RGB samples~\cite{chen2020oasis,Chen17Snow}. This labeling process is expensive and time-consuming, and crucially does not allow regenerating the image dataset. In Sec.~\ref{sec:soundness}, we consider our pipeline vs. OASIS, one of the largest and most diverse of these benchmarks, and demonstrate that models trained on our starter dataset already match human-level performance on OASIS---outperforming the same architectures models trained on OASIS itself.

\textbf{Vision-Focused Simulators.} Like our platform, simulators typically take a textured mesh as the representation of the scene and aim to produce realistic sensory inputs~\cite{habitat19iccv,gibson}. While spiritually similar to the pipeline proposed in this paper, the current generation of simulators is designed first and foremost to train embodied agents. They prioritize rendering speed and real-time mechanics at the cost of photorealism and cue diversity~\cite{vizdoom,mnih-dqn-2015}. Extending such simulators to handle additional cues or to parametrically render out vision datasets often requires writing new components of the simulator codebase (usually in C++, CUDA, or OpenGL), a surmountable but unpleasant barrier to entry. In contrast, our platform extends Blender which ``supports the entirety of the 3D pipeline''~\cite{community2018blender} and provides Python bindings that will be intuitive to most vision researchers, and we implement many of these cues and sampling methods out-of-the-box. In short, we provide a bridge between simulators and static vision datasets.

\textbf{Multi-Task Datasets.}
Vision-based multi-task learning (MTL), like computer vision in general, shows a general bias towards recognition. MTL datasets often take different shades of classification as the core problem of interest~\cite{lake2019omniglot, VinyalsBLKW16Matching, maltoni2018icifar}. In particular, MTL literature often focuses on binary attribute classification in specialized domains, such as Caltech-UCSD Birds~\cite{WelinderEtal2010} or CelebA~\cite{liu2015faceattributes}. Visual MTL datasets that contain non-recognition tasks often contain only a single domain or a few tasks (NYU~\cite{silberman2012indoor}, CityScapes~\cite{cordts2016cityscapes} or Taskonomy~\cite{taskonomy2018}).
Sometimes, MTL papers take mix datasets for a ``single'' task and consider each dataset as a different task~\cite{liu2019end,ranftl2019towards,lambert2020mseg,Ranftl2021}.

In general, the multi-task learning literature has not converged on a standardized definition of the setting or dataset. Recent work has demonstrated that MTL methods developed on existing datasets seem to specialize to their respective developement set and do not perform well on large, realistic datasets, or on other tasks~\cite{vandenhende2019branched,vandenhende2021multi, zhang2019sidetuning}. This underscores the importance of developing \emph{realistic training setting and datasets} that generalizes to real-world scenarios.

\textbf{Data Augmentation + Domain Randomization.} 
Data augmentation is a way to modify the data or training regimen so that the trained model exhibits desirable invariances (or equivariances).
During training, \emph{any} transformation of sensor inputs that determines a unique (possibly identity) transformation on the label can be used as ``augmented'' data. For example, simple 2D augmentations such as 2D affine transformations, crops, and color changes that leave the labels unchanged are the common in computer vision~\cite{chen2020simple,grill2020bootstrap}, since they can be used even when datasets lack 3D geometry information.
In robotics and reinforcement learning where 3D simulators are more standard, data augmentation was introduced as ``domain randomization''~\cite{tobin2017domain}, and common augmentations include texture and background randomization on the scene mesh. Recently, \cite{denninger2019blenderproc} introduced a Blender-based approach for doing domain randomization and creating static datasets of RGB, depth, and surface normals from SunCG~\cite{sun2019deep}. 

Our pipeline makes all these augmentations available for static computer vision datasets: not just flips/crops/texture randomization, but also dense viewpoints, multi-view consistency, Euclidean transforms, lens flare, etc.). We implement and examine depth-of-field augmentation in Sec.~\ref{sec:refocus}.

\textbf{Auto Labeling} is an umbrella term for a group of data labeling procedures that harness structure in the data as constraints in order to prune or propagate labels and save annotation labor. This is accomplished primarily by I) pre-trained models as noisy annotator (e.g. ~\cite{armeni_iccv19, hinton2015distilling, selftraining}),  and/or II) aggregating and filtering annotations based on known constraints (e.g. backprojection error, bundle adjustment, temporal smoothness, or ~\cite{mirrorflow, leftRightConsistency, 2d3ds}).
Our pipeline has connections to auto labeling in the sense that we make use of the strong structure present in 3D scanned data to compute and propagate labels across images and reduce the load of (automatically or manually) labeling each image.

\section{Pipeline Overview}




We call our pipeline \handle as it aspires to encapsulate comprehensive scene information (``omni'') in the generated ``data''. \noindent\href{https://omnidata.vision}{\textbf{Try a live example here}} to get acquainted with the pipeline. The example uses the CLI and a YAML-like config file to generate images from a textured mesh in Replica~\cite{replica19arxiv}.


\noindent\textbf{Inputs:}
The annotator operates upon the following inputs:
\begin{itemize}[leftmargin=4mm,itemsep=-0.2em]
    \vspace{-2mm}
    \item Untextured Mesh (.obj or .ply)
    \vspace{-1mm}
    \item {\emph{Either:} Mesh Texture or Aligned RGB Images}
    \vspace{-1mm}
    \item {\emph{Optional:} Pre-Generated Camera Pose File}
    \vspace{-2mm}
\end{itemize}
A 3D pointcloud can be used as well: simply mesh the pointcloud using a standard mesher like COLMAP~\cite{schoenberger2016sfm}. An example of meshing and using a 3D pointcloud with the annotator, as well as a more complete description of inputs are available in the \href{http://omnidata.vision/supplementary_material/}{supplementary}.

\noindent\textbf{Outputs:}
The pipeline generates 21 mid-level cues in the initial release. All labels are available for all generated images (or videos). Fig.~\ref{fig:pull} provides a visual summary of the different types of outputs. A detailed description of the default mid-level cues and additional outputs provided by the \handle annotator is included in the \href{http://omnidata.vision/supplementary_material/}{supplementary}.

\subsection{Sampling and Generation}\label{sec:generation_pipeline}

\begin{figure}[H]
    \vspace{-2.0mm}	
    \centering
	\includegraphics[width=0.9\linewidth]{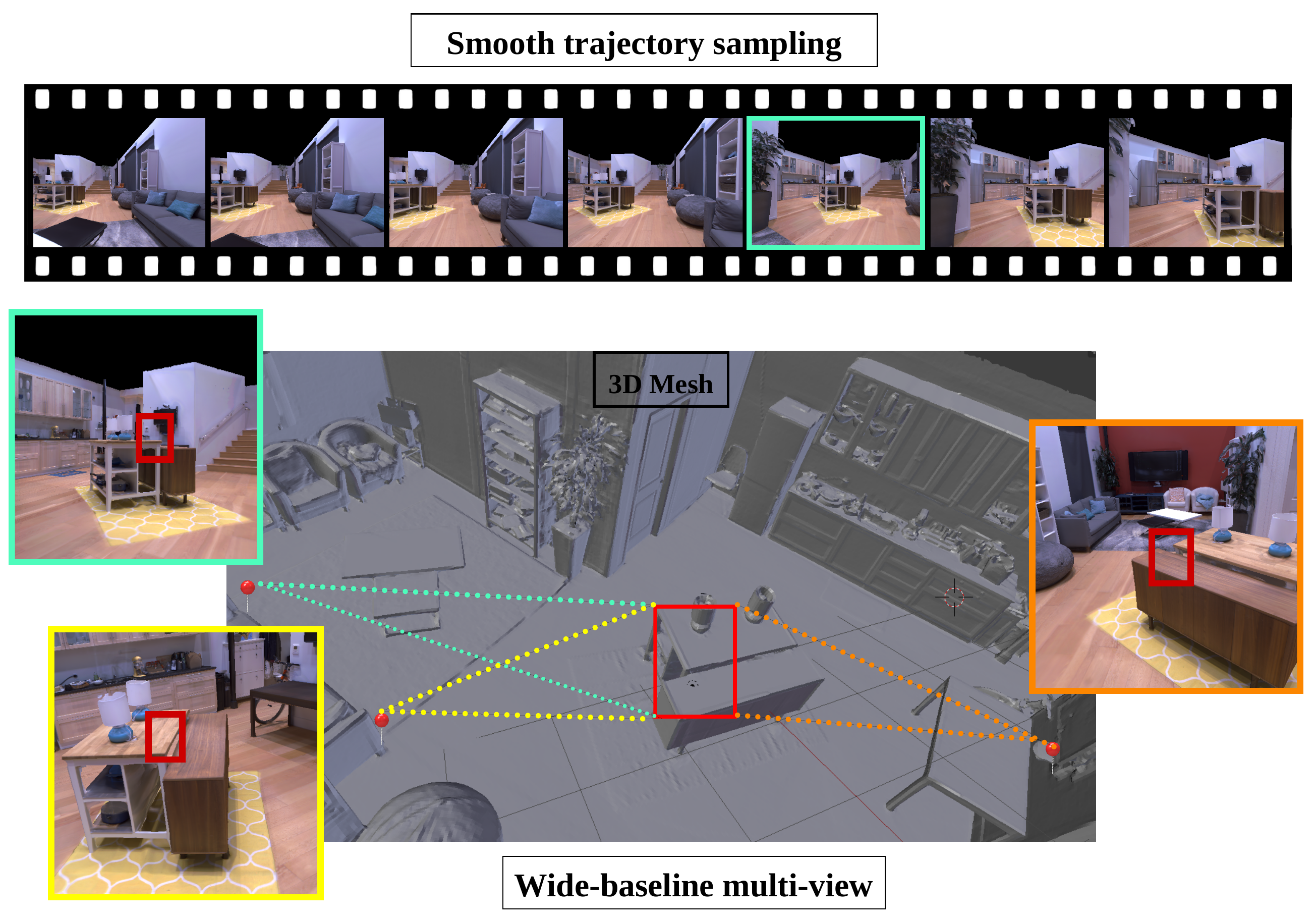}
	\caption{\footnotesize{\textbf{Wide- and narrow-baseline dense view sampling.} Each point-of-interest can be viewed by a guaranteed minimum number of cameras. We also provide an option for creating denser views with narrower baselines (\eg similar to consecutive video frames) that is crucial for inverse rendering methods.}.\vspace{-2mm}}
	\label{fig:view_sampling}
\end{figure}

\begin{figure}[H]
		\vspace{-5mm}
		\centering
		\includegraphics[width=0.9\columnwidth]{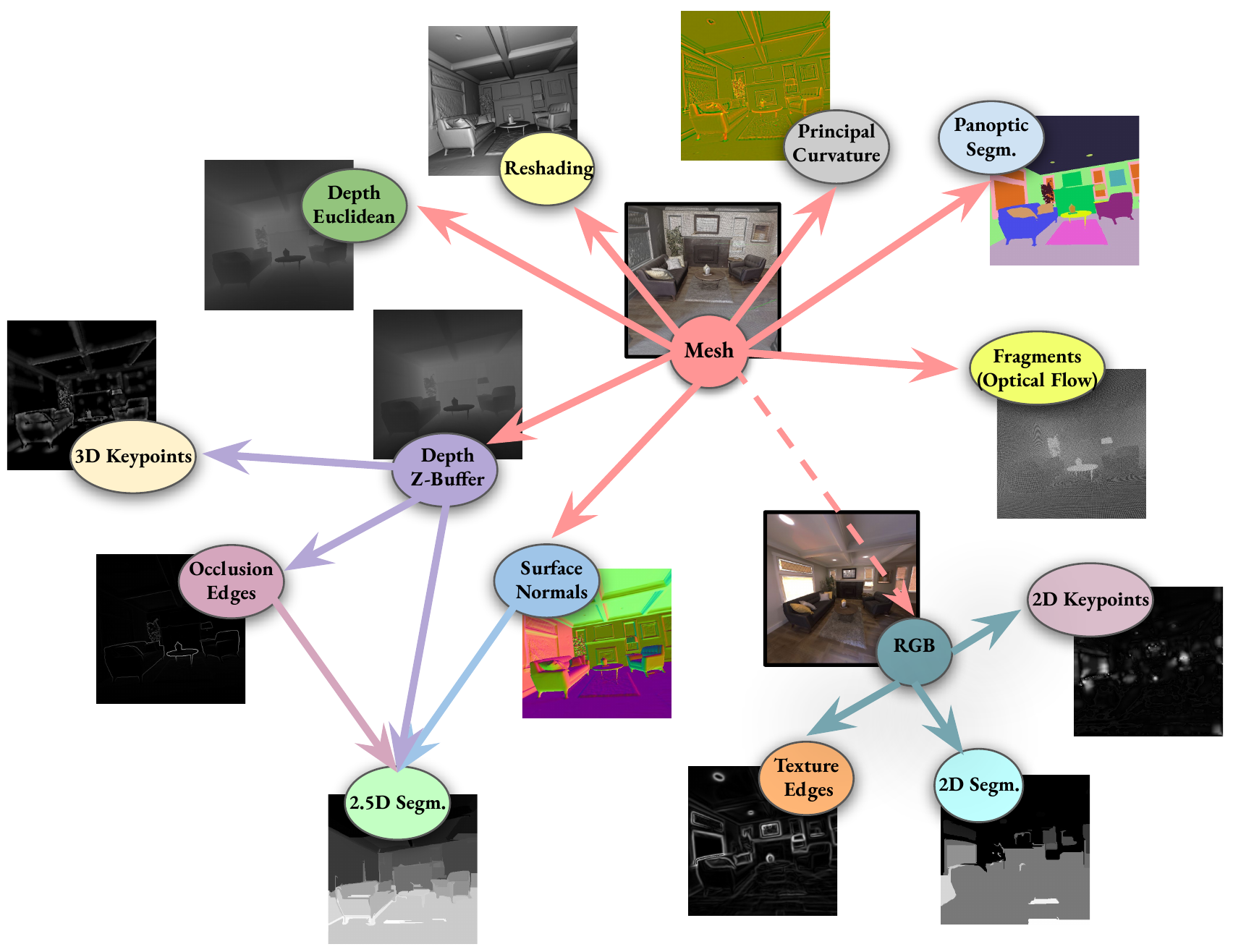}
	    \vspace{-2.5mm}
		\caption{\footnotesize{\textbf{DAG of processing pipeline.} The pipeline uses some of the mid-level cues to produce others. The ordering of this process (for image-like cues) is shown by the DAG.}}
			\vspace{-3mm}
		\label{fig:processing_dag}
	\end{figure}

\noindent In this section we provide a high-level outline of the generation and rendering process (see Fig.~\ref{fig:methodology}), deferring full details to the \href{http://omnidata.vision/supplementary_material/}{supplementary}. 
\vspace{-1mm}
\begin{description}[leftmargin=2mm,itemsep=-0.2em]
    \vspace{-2mm}
    \item \hspace{-1.7mm}\textbf{First}, the annotator generates camera locations (Fig.~\ref{fig:methodology} II) and points-of-interest (Fig.~\ref{fig:methodology} III) along the mesh.
    \vspace{-1mm}
    \item \hspace{-1.7mm}\textbf{Second}, for each camera and each point-of-interest, it creates a view from that camera fixated on the point (three fixated views are depicted in the lower part of Fig.~\ref{fig:view_sampling}).
    \vspace{-1mm}
    \item \hspace{-1.7mm}\textbf{Third}, for each space-point-view triplet, the annotator renders (Fig.~\ref{fig:processing_dag}) all the mid-level cues (Fig.~\ref{fig:methodology} IV).
    \vspace{-3mm}
\end{description}
Each step is elaborated next.

\begin{figure*}
		\vspace{-2mm}
		\centering
		\includegraphics[width=0.95\textwidth]{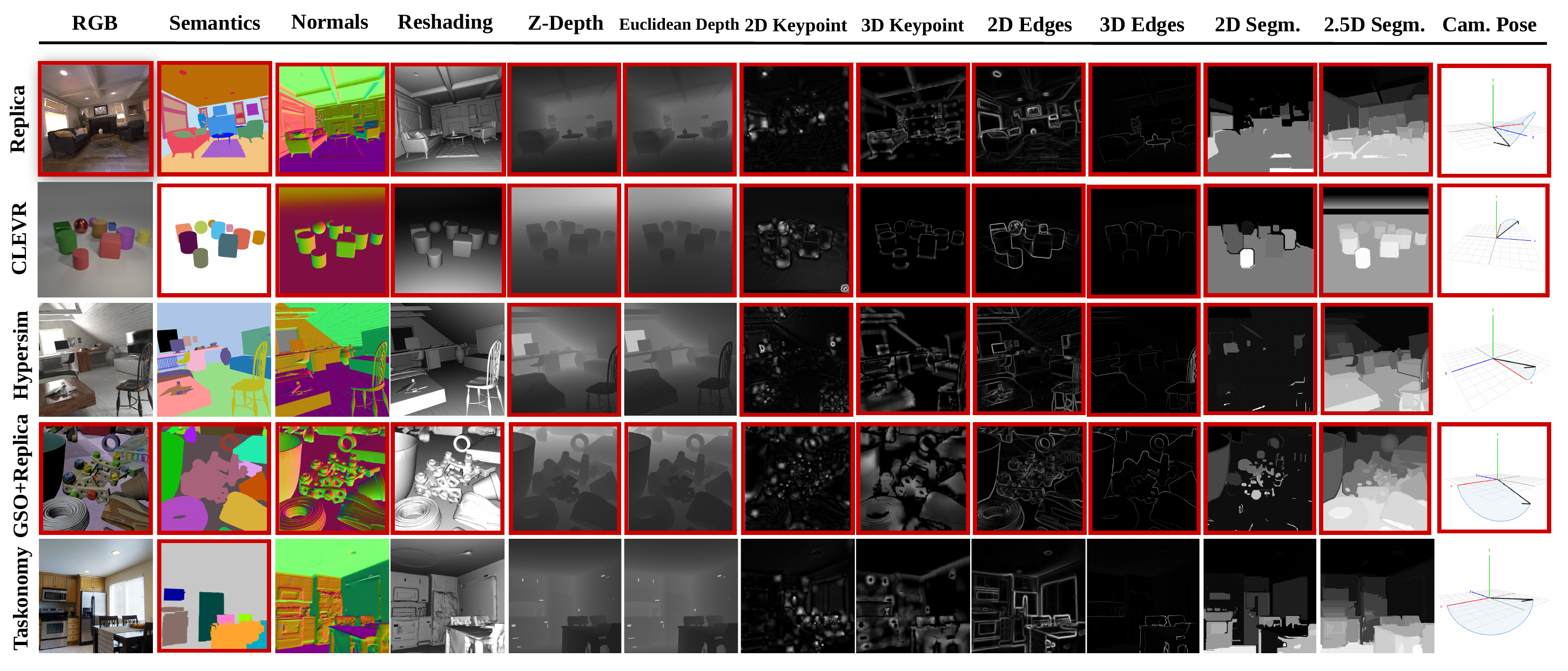}
		\vspace{-2.5mm}
		\caption{\footnotesize{\textbf{Mid-level cues provided for the starter set.} 12 out of 21 mid-level cues visualized for each component dataset in the starter set, which contains both scenes and objects. Images with \textcolor{red}{red} borders indicate cues that were not included in the original data. Fig.~\ref{fig:pull} visualizes all 21 cues.}}
			\vspace{-2mm}
		\label{fig:starter_dataset_modalities}
	\end{figure*}

\noindent\textbf{Camera and Point Sampling:}
Camera locations can be provided (if the mesh comes with aligned RGB), or as in Fig.~\ref{fig:methodology} II, the annotator generates cameras in each space so that cameras are not inside or overlapping with the mesh (default: cameras generated via Poisson-disc sampling to cover the space). Points-of-interest are then sampled from the mesh with a user-specified sampling strategy (default: uniform sampling of each mesh face and then uniform sampling on that face). Cameras and points are then filtered so that each camera sees at least one point and each point is seen by at least some user-specified minimum number of cameras (default: 3).

\noindent\textbf{View Sampling:}
The annotator provides two default methods for generating views of each point. The first method (wide-baseline) generates images while the second, (smooth-trajectory mode) generates videos.

\vspace{-2mm}
\begin{itemize}[leftmargin=4mm]\setlength\itemsep{-0.4em}
    \item \emph{Wide-baseline multi-view:} A view is saved for each space-camera-point combination where there is an unobstructed line-of-sight between the camera center and the point-of-interest. The camera is fixated on the point-of-interest, as shown in Fig.~\ref{fig:view_sampling}, bottom.
    \item \emph{Smooth trajectory sampling:} For each point of interest, a subset of cameras with a fixated view of the point are selected, and a smooth cubic-spline trajectory is interpolated between the cameras. Views of the point are generated for cameras at regular intervals along this trajectory (see Fig.~\ref{fig:view_sampling}, top).
\end{itemize}

\noindent\textbf{Rendering mid-level cues:} Since no single piece of software was able to provide all mid-level cues, we created an interconnected pipeline connecting several different pieces of software that are all freely available and open-source. We tried to primarily use Blender (a 3D creation suite), since it has an active user and maintenance community, excellent documentation, and python bindings for almost everything. Used by professional animators and artists, it is is generally well-optimized. The overall pipeline is fairly complex, so we defer a full description to the supplementary. The order of cue generation is shown in Fig.~\ref{fig:processing_dag}. The full code is available \href{https://omnidata.vision}{on our website}.

\noindent\textbf{Performance:} The annotator generates labels in any resolution. Each space+point+view+cue label in the starter dataset ($512\times512$) typically takes 1-4 seconds on server or desktop CPUs and can be parallelized over many machines.

\subsection{Ecosystem Tools}
\noindent
To simplify adoption, the following tools are available \href{http://omnidata.vision}{on our website} and the associated GitHub repository:

\vspace{-0.5em}
\begin{description}[leftmargin=3.7mm]\setlength\itemsep{-0.35em}
    \item
        \hspace{-2.2mm}
        \textbf{Pipeline code} and documentation.
    \item 
        \hspace{-2.2mm}
        \textbf{Docker} containing the annotator and properly linked software (Blender~\cite{community2018blender}, compatible Python versions, MeshLab~\cite{cignoni2008meshlab},  etc.).
    \item
        \hspace{-2.2mm}
        \textbf{Dataloaders} in PyTorch for correctly and efficiently loading the resulting dataset
    \item
        \hspace{-2.2mm}
        \textbf{Starter dataset} of 14.5 million images with associated labels for each task
    \item
        \hspace{-2.2mm}
        \textbf{Convenience utilities} for downloading and manipulating data and  \emph{automatically filtering} misaligned meshes (description and sensitivity analysis in the \href{http://omnidata.vision/supplementary_material/}{supplementary}).
    \item
        \hspace{-2.2mm}
        \textbf{Pretrained models and code}, including the first publicly available implementation of MiDaS~\cite{ranftl2019towards} training code.
\end{description}
\vspace{-0.5em}


% ...


\section{Starter Dataset Overview}
\label{sec:starter}

We provide a relatively large starter dataset of data annotated with the \handle annotator. The dataset comprises roughly \textbf{14.5 million} images from scenes that are both scene- and object-centric. Fig.~\ref{fig:starter_dataset_modalities} shows sample images from the starter dataset along with 12 of the 21 mid-level cues provided. Cues that are not present in the original dataset are indicated with a \textcolor{red}{red} border. We evaluate this starter dataset on existing benchmarks in Sec.~\ref{sec:soundness}. Note that the dataset could be straightforwardly extended to other existing outdoor and driving datasets such as GTA5~\cite{Richter_2016_ECCV}, CARLA~\cite{Dosovitskiy17}, or Tanks and Temples~\cite{knapitsch2017tanks}.

\begin{figure*}
		\vspace{-7mm}
		\centering

		\includegraphics[width=1\textwidth]{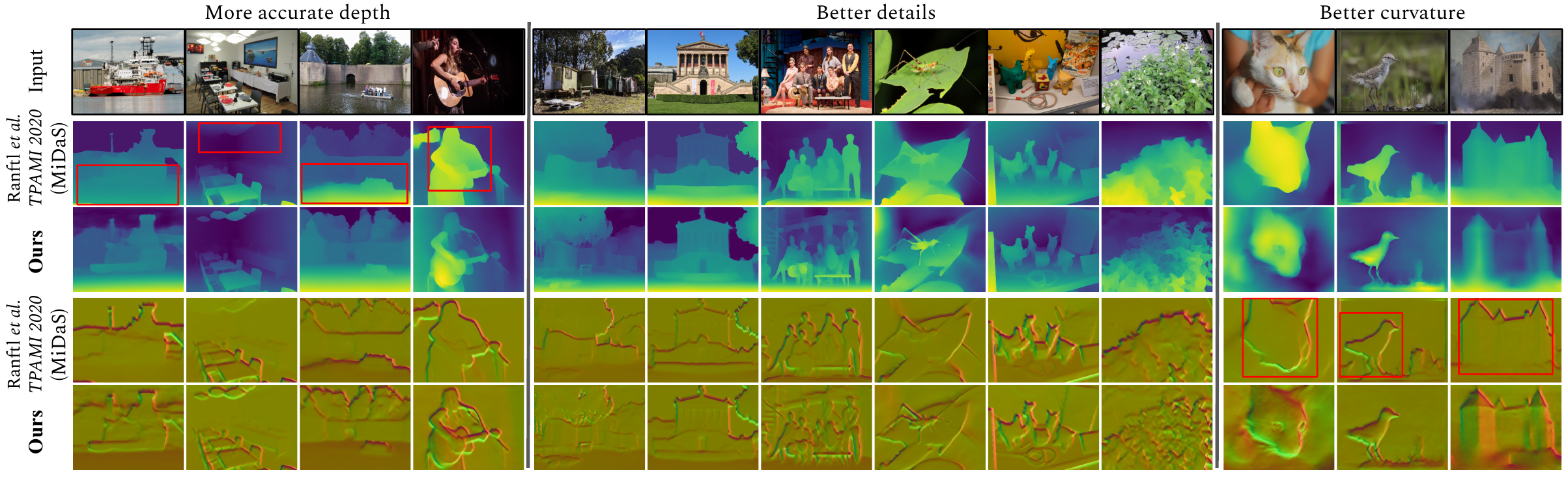}
		\vspace{-6mm}
		\caption{\footnotesize{\textbf{Qualitative comparison with MiDaS on zero-shot OASIS depth estimation.} The last 2 rows show the surface normals extracted from the depth predictions. Our model predicts more accurate depth (left), and also outperforms in recovering the fine-grained details (middle). As shown by the extracted surface normal in the last 3 columns (right), our depth predictions better reflect the curvature and true shape of the objects, while the same regions appear flat in the predictions by the MiDaS model. The red rectangles highlight the regions useful for comparison [best viewed zoomed in]. }}
		\vspace{-3mm}
		\label{fig:depth-oasis}
	\end{figure*}


\subsection{Datasets Included}
\noindent The starter data was created from 7 mesh-based datasets:

\noindent\textbf{Indoor scene datasets:} Replica~\cite{replica19arxiv}, HyperSim~\cite{roberts:2020}, Taskonomy~\cite{taskonomy2018}, Habitat-Matterport (HM3D)

\noindent\textbf{Aerial/outdoor datasets:} BlendedMVG~\cite{yao2020blendedmvs}

\noindent\textbf{Diagnostic/Structured datasets:} CLEVR~\cite{johnson2017clevr}

\noindent\textbf{Object-centric datsets:} To provide object-centric views in addition to scene-centric ones, we create a dataset of Google Scanned Objects~\cite{gsoignition} scattered around buildings from the Replica~\cite{replica19arxiv} dataset (similar to how ObjectNet~\cite{barbu2019objectnet} diversified images for image classification). We used the Habitat~\cite{habitat19iccv} environment to generate physically plausible scenes, and generated different densities of objects. Examples of images are shown in Fig.~\ref{fig:starter_dataset_modalities}, and a full description of the generation process is available in the supplementary.

\subsection{Dataset Statistics}
The starter dataset contains \textbf{14,601,449} images from \textbf{2,414} spaces. Views are both scene- and object-centric, and they are labeled with each modality listed in Fig.\ref{fig:pull}. Camera field-of-view is sampled from a truncated normal distribution between $30^{\circ}$ and $125^{\circ}$ with mean $77.5^{\circ}$, and camera roll is uniform in $[-10^{\circ}, 10^{\circ}]$. Tab.~\ref{table:dataset_statistics} contains data broken down to sub-datasets.

\setlength{\tabcolsep}{7pt}
\renewcommand{\arraystretch}{1}
\begin{table}[htb]
		\centering
		\resizebox{\linewidth}{!}{
		\begin{tabular}{|lll|lll|ccc|c|}
			\hline
			\multicolumn{3}{|c|}{} & \multicolumn{3}{c|}{Images} & \multicolumn{3}{c|}{Spaces} & \multicolumn{1}{c|}{Points} \\
			\multicolumn{3}{|l|}{Dataset} & \multicolumn{1}{c}{Train} & \multicolumn{1}{c}{Val} & \multicolumn{1}{c|}{Test} & \multicolumn{1}{c}{Train} & 
			\multicolumn{1}{c}{Val} & 
			\multicolumn{1}{c|}{Test} & 
			\multicolumn{1}{c|}{}  \\ 
			
			\hline
			\hline
			
			\multicolumn{3}{|l|}{CLEVR} & \multicolumn{1}{c}{60,000} &
			\multicolumn{1}{c}{6,000} &
			\multicolumn{1}{c|}{6,000} &
			\multicolumn{1}{c}{1}  & 
			\multicolumn{1}{c}{0} & 
			\multicolumn{1}{c|}{0} & 
			\multicolumn{1}{c|}{72,000}  \\
			
			\multicolumn{3}{|l|}{Replica} & \multicolumn{1}{c}{56,783} &
			\multicolumn{1}{c}{23,725} &
			\multicolumn{1}{c|}{23,889} &
			\multicolumn{1}{c}{10}  & 
			\multicolumn{1}{c}{4} & 
			\multicolumn{1}{c|}{4} & 
			\multicolumn{1}{c|}{4,150}  \\
			
			\multicolumn{3}{|l|}{Replica + GSO} & \multicolumn{1}{c}{107,404} &
			\multicolumn{1}{c}{43,450} &
			\multicolumn{1}{c|}{42,665} &
			\multicolumn{1}{c}{10}  & 
			\multicolumn{1}{c}{4} & 
			\multicolumn{1}{c|}{4} & 
			\multicolumn{1}{c|}{31,167}  \\
			
			\multicolumn{3}{|l|}{Hypersim} & \multicolumn{1}{c}{59,543} &
			\multicolumn{1}{c}{7,386} &
			\multicolumn{1}{c|}{7,690} &
			\multicolumn{1}{c}{365}  & 
			\multicolumn{1}{c}{46} & 
			\multicolumn{1}{c|}{46} & 
			\multicolumn{1}{c|}{74,619}  \\
			
			\multicolumn{3}{|l|}{Taskonomy} & \multicolumn{1}{c}{3,416,314} &
			\multicolumn{1}{c}{538,567} &
			\multicolumn{1}{c|}{629,581} &
			\multicolumn{1}{c}{379}  & 
			\multicolumn{1}{c}{75} & 
			\multicolumn{1}{c|}{79} & 
			\multicolumn{1}{c|}{684,052}  \\
			
			\multicolumn{3}{|l|}{BlendedMVG} & \multicolumn{1}{c}{79,023} &
			\multicolumn{1}{c}{16,787} &
			\multicolumn{1}{c|}{16,766} &
			\multicolumn{1}{c}{341}  & 
			\multicolumn{1}{c}{74} & 
			\multicolumn{1}{c|}{73} & 
			\multicolumn{1}{c|}{112,576}  \\
			
			\multicolumn{3}{|l|}{Habitat-Matterport} & \multicolumn{1}{c}{8,470,855} &
			\multicolumn{1}{c}{1,061,021} &
			\multicolumn{1}{c|}{-} &
			\multicolumn{1}{c}{800}  & 
			\multicolumn{1}{c}{100} & 
			\multicolumn{1}{c|}{-} & 
			\multicolumn{1}{c|}{564,328}  \\

			\hline
			
			\multicolumn{3}{|l|}{Total (no CLEVR)} & \multicolumn{1}{c}{12,189,922} &
			\multicolumn{1}{c}{1,690,936} &
			\multicolumn{1}{c|}{720,591} &
			\multicolumn{1}{c}{1,905}  & 
			\multicolumn{1}{c}{303} & 
			\multicolumn{1}{c|}{206} & 
			\multicolumn{1}{c|}{1,434,892}  \\

			\hline
		\end{tabular}
		}
		\vspace{-2.5mm}
		\captionof{table}{\footnotesize{\textbf{Component dataset statistics.} Breakdown of train/val/test split sizes in each of the components of the starter datset. }}\label{table:dataset_statistics}
		\vspace{-3mm}
			
\end{table}
\setlength{\tabcolsep}{6pt}
\renewcommand{\arraystretch}{1}


\subsection{Soundness for Existing Computer Vision}
\label{sec:soundness}
We demonstrate that the generated dataset is capable of training standard, modern vision systems to state-of-the-art performance on existing benchmarks.
Once we have established that the models can be trusted, we further provide a few transfer experiments to quantify how related the different component datasets are. 

We show that the models trained on a 5-dataset portion of the starter dataset (4M images) for depth and surface normal estimation have state-of-the-art performance on the in-the-wild OASIS benchmark.
To demonstrate the effectiveness of the pipeline for semantic tasks, we show that the predictions from a network trained for panoptic segmentation on a smaller 3-dataset portion (1M images) are of similar quality to models trained on COCO~\cite{MSCoco}. Full experimental details and more results are available in the \href{http://omnidata.vision/supplementary_material/}{supplementary}.

\setlength{\tabcolsep}{8pt}
\renewcommand{\arraystretch}{1.2}
\begin{table}[ht]
		\centering
		\vspace{-1mm}
		\resizebox{\linewidth}{!}{
		\begin{tabular}{llllll}
			\hline
			
			\multicolumn{1}{|c|}{Method} &
			\multicolumn{1}{c|}{Test Data} &
			\multicolumn{1}{c}{L1 Error ($\downarrow$)} &
			\multicolumn{1}{c}{$\delta>1.25$ ($\downarrow$)} &
			\multicolumn{1}{c}{$\delta>1.25^2$ ($\downarrow$)} &
			\multicolumn{1}{c|}{$\delta>1.25^3$ ($\downarrow$)} \\
			\hline
			\hline
			
			\multicolumn{1}{|c|}{XTC~\cite{zamir2020consistency}} &
			\multicolumn{1}{c|}{} &
			\multicolumn{1}{c}{1.180} &
			\multicolumn{1}{c}{85.28} &
			\multicolumn{1}{c}{71.86} &
			\multicolumn{1}{c|}{60.22} \\ 
			
			\multicolumn{1}{|c|}{MiDaSv3~\cite{Ranftl2021}} &
			\multicolumn{1}{c|}{OASIS~\cite{chen2020oasis}} &
			\multicolumn{1}{c}{0.8057} &
			\multicolumn{1}{c}{82.03} &
			\multicolumn{1}{c}{67.25} &
			\multicolumn{1}{c|}{55.35} \\ 
			
			\multicolumn{1}{|c|}{\textbf{\handle}} &
			\multicolumn{1}{c|}{} &
			\multicolumn{1}{c}{\textbf{0.7901}} &
			\multicolumn{1}{c}{\textbf{81.00}} &
			\multicolumn{1}{c}{\textbf{65.22}} &
			\multicolumn{1}{c|}{\textbf{52.93}} \\
			
			\hline
			
			\multicolumn{1}{|c|}{XTC~\cite{zamir2020consistency}} &
			\multicolumn{1}{c|}{} &
			\multicolumn{1}{c}{0.5279} &
			\multicolumn{1}{c}{70.41} &
			\multicolumn{1}{c}{49.90} &
			\multicolumn{1}{c|}{36.28} \\ 
			
			\multicolumn{1}{|c|}{MiDaSv3~\cite{Ranftl2021}} &
			\multicolumn{1}{c|}{NYU~\cite{silberman2012indoor}} &
			\multicolumn{1}{c}{0.3838} &
			\multicolumn{1}{c}{63.84} &
			\multicolumn{1}{c}{41.65} &
			\multicolumn{1}{c|}{28.97} \\ 
			
			\multicolumn{1}{|c|}{\textbf{\handle}} &
			\multicolumn{1}{c|}{}&
			\multicolumn{1}{c}{\textbf{0.2878}} &
			\multicolumn{1}{c}{\textbf{51.73}} &
			\multicolumn{1}{c}{\textbf{30.98}} &
			\multicolumn{1}{c|}{\textbf{20.86}} \\

			\hline
		\end{tabular}
		}
        \vspace{-2.5mm}
		\captionof{table}{\footnotesize{\textbf{Zero-shot depth estimation.} On NYU and OASIS, a DPT-Hybrid trained on the \handle starter dataset is comparable or better then the same model trained on existing depth datasets.
		}}\label{table:depth-oasis}
    \vspace{-2mm}
\end{table}

\setlength{\tabcolsep}{6pt}
\renewcommand{\arraystretch}{1}

\vspace{-1mm}
\renewcommand{\arraystretch}{1.1}
\setlength{\tabcolsep}{5pt}
\begin{table}[H]
		\centering
		\resizebox{\linewidth}{!}{
		\begin{tabular}{|l|l|ll|lll|cc|}
			\hline
			\multicolumn{1}{|c|}{} &
			\multicolumn{1}{c|}{} & \multicolumn{2}{c|}{Anglular Error$^\circ$} & \multicolumn{3}{c|}{$\%$ Within $t^{\circ}$} &
			\multicolumn{2}{c|}{Relative Normal}
			\\ 
			\multicolumn{1}{|c|}{Method} & \multicolumn{1}{c|}{Training Data} & \multicolumn{1}{c}{Mean} &
			\multicolumn{1}{c|}{Median} &
			\multicolumn{1}{c}{$11.25^{\circ}$} &
			\multicolumn{1}{c}{$22.5^{\circ}$} &
			\multicolumn{1}{c|}{$30^{\circ}$} &
			\multicolumn{1}{c}{$AUC_{o}$} &
			\multicolumn{1}{c|}{$AUC_{p}$} \\
			
			\hline
			\hline
			
			\multicolumn{1}{|c|}{Hourglass~\cite{Chen16MonocularDepth}} &
			\multicolumn{1}{c|}{OASIS~\cite{chen2020oasis}} &
			\multicolumn{1}{c}{\textbf{23.91}} &
			\multicolumn{1}{c|}{18.16} &
			\multicolumn{1}{c}{\textbf{31.23}} &
			\multicolumn{1}{c}{59.45}  & 
			\multicolumn{1}{c|}{\textbf{71.77}} & 
			\multicolumn{1}{c}{0.5913} &
			\multicolumn{1}{c|}{0.5786} \\

            \multicolumn{1}{|c|}{Hourglass~\cite{Chen16MonocularDepth}} &
			\multicolumn{1}{c|}{SNOW~\cite{Chen17Snow}} &
			\multicolumn{1}{c}{31.35} &
			\multicolumn{1}{c|}{26.97} &
			\multicolumn{1}{c}{13.98} &
			\multicolumn{1}{c}{40.20}  & 
			\multicolumn{1}{c|}{56.03} & 
			\multicolumn{1}{c}{0.5329} &
			\multicolumn{1}{c|}{0.5016} \\
			
            \multicolumn{1}{|c|}{Hourglass~\cite{Chen16MonocularDepth}} &
			\multicolumn{1}{c|}{NYU~\cite{Silberman2012}} &
			\multicolumn{1}{c}{35.32} &
			\multicolumn{1}{c|}{29.21} &
			\multicolumn{1}{c}{14.23} &
			\multicolumn{1}{c}{37.72}  & 
			\multicolumn{1}{c|}{51.31} & 
			\multicolumn{1}{c}{0.5467} &
			\multicolumn{1}{c|}{0.5132} \\
			
            \multicolumn{1}{|c|}{PBRS~\cite{Zhang16PBRS}} &
			\multicolumn{1}{c|}{NYU~\cite{Silberman2012}} &
			\multicolumn{1}{c}{38.29} &
			\multicolumn{1}{c|}{33.16} &
			\multicolumn{1}{c}{11.59} &
			\multicolumn{1}{c}{32.14}  & 
			\multicolumn{1}{c|}{45.00} & 
			\multicolumn{1}{c}{0.5669} &
			\multicolumn{1}{c|}{0.5253} \\
			
			\multicolumn{1}{|c|}{UNet~\cite{Ronneberger15Unet}} &
			\multicolumn{1}{c|}{SunCG~\cite{song2016ssc}} &
			\multicolumn{1}{c}{35.42} &
			\multicolumn{1}{c|}{28.70} &
			\multicolumn{1}{c}{12.31} &
			\multicolumn{1}{c}{38.51}  & 
			\multicolumn{1}{c|}{52.15} & 
			\multicolumn{1}{c}{0.5871} &
			\multicolumn{1}{c|}{0.5318} \\
			\hline
			
            \multicolumn{1}{|c|}{\textbf{UNet}~\cite{Ronneberger15Unet}} &
			\multicolumn{1}{c|}{\textbf{\handle}} &
			\multicolumn{1}{c}{24.87} &
			\multicolumn{1}{c|}{\textbf{18.04}} &
			\multicolumn{1}{c}{31.02} &
			\multicolumn{1}{c}{\textbf{59.53}}  & 
			\multicolumn{1}{c|}{71.37} & 
			\multicolumn{1}{c}{\textbf{0.6692}} &
			\multicolumn{1}{c|}{\textbf{0.6758}} \\
			
			
            \multicolumn{1}{|c|}{Human (Approx.)} &
			\multicolumn{1}{c|}{-} &
			\multicolumn{1}{c}{17.27} &
			\multicolumn{1}{c|}{12.92} &
			\multicolumn{1}{c}{44.36} &
			\multicolumn{1}{c}{76.16}  & 
			\multicolumn{1}{c|}{85.24} & 
			\multicolumn{1}{c}{0.8826} &
			\multicolumn{1}{c|}{0.6514} \\

			\hline
		\end{tabular}
		}
		\vspace{-2mm}
		\captionof{table}{\footnotesize{\textbf{Zero-shot surface normal estimation on OASIS.} A UNet trained on the \handle starter dataset matched or outperformed models trained on OASIS itself, and it matched human-level $AUC_{p}$. Notice that the first row is not zero-shot since it's trained on OASIS.
		}}\label{table:oasis_results}
    \addtolength{\tabcolsep}{4pt} 
    \vspace{-4mm}
\end{table}
\setlength{\tabcolsep}{6pt}
\renewcommand{\arraystretch}{1}

\begin{figure*}[t!]
		\vspace{-2mm}
			\hspace{4mm}
		\centering
		\includegraphics[width=1.0\textwidth]{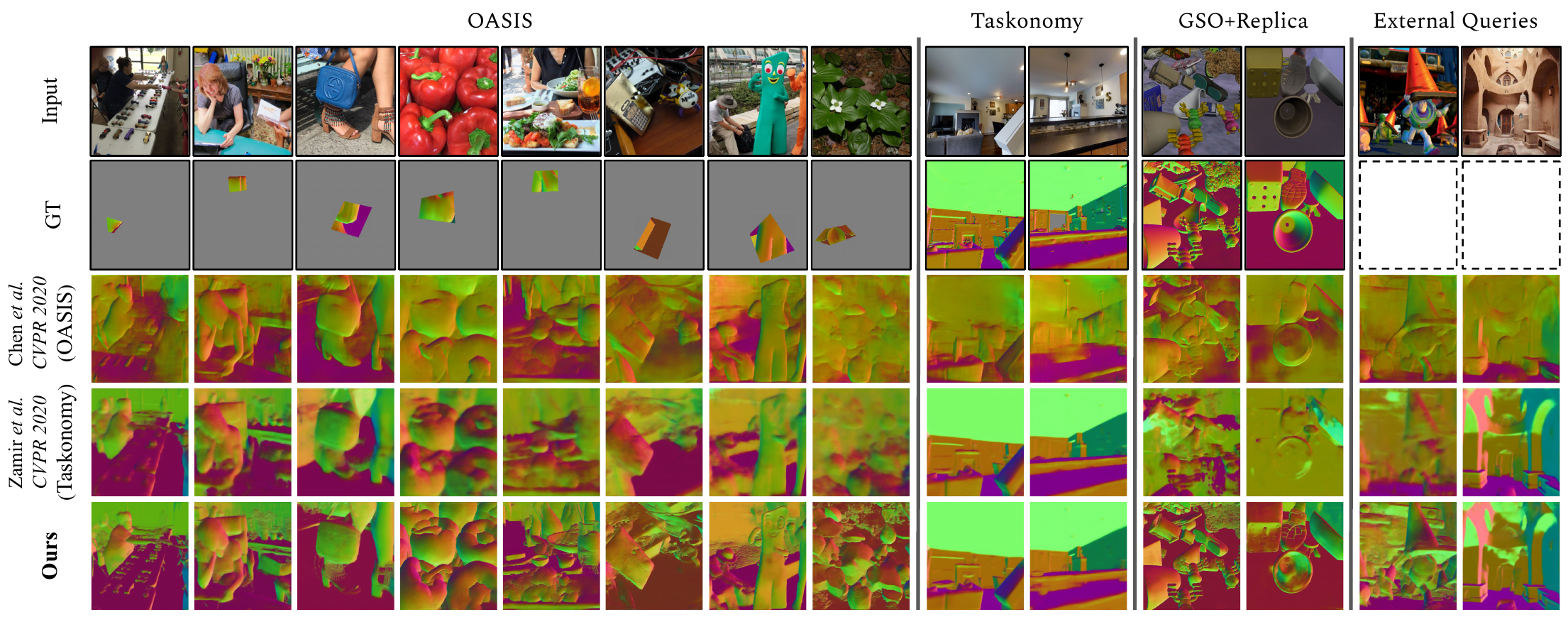}
		\vspace{-6mm}
		\caption{\footnotesize{\textbf{Qualitative results of zero-shot surface normal estimation.} The 3 models are trained on OASIS\cite{chen2020oasis}, Full Taskonomy\cite{zamir2020consistency,taskonomy2018}, and our starter set. Queries are from 3 different datasets (OASIS, Taskonomy, GSO+Replica) in addition to some external queries in the last 2 columns (no ground truth available) which show the generalization of the models to external data [best viewed zoomed in].}}
			\vspace{-3mm}
		\label{fig:surface_normals}
	\end{figure*}

\noindent\textbf{Monocular depth estimation:} The current best approach for depth estimation is to aggregate multiple smaller datasets and train with scale- and shift-invariant losses~\cite{ranftl2019towards,Ranftl2021} to handle the different unknown depth ranges and scales. As of this writing, the DPT-based~\cite{Ranftl2021} models from ``MiDaS v3.0''~\cite{Ranftl2021} define the state-of-the-art, especially on NYU~\cite{silberman2012indoor}. We adopt a similar setting to MiDaS v3.0, but train on a 5-dataset portion of our starter dataset instead of their 10-dataset mix\footnote{
MiDaS v3.0 also uses MTAN~\cite{liu2019end} for dataset balancing, and though in Sec.~\ref{sec:mtl_exp} we examine MTAN (it indeed helped on our dataset), we used here a naive sampling strategy in order to be consistent with the majority of the other models in this paper.
}.

As in \cite{Ranftl2021}, we evaluate zero-shot cross-dataset transfer with test predictions and GT aligned in scale and shift in inverse-depth space.
Tab.~\ref{table:depth-oasis} shows that the DPT-Hybrid trained on our starter dataset outperforms MiDaS DPT-Hybrid on both the test set of NYU~\cite{silberman2012indoor} and the validation split of OASIS (the test GT is not available).
The error metrics use $\delta=max(\frac{d}{d^*}, \frac{d^*}{d})$ where $d$ and $d^*$ are aligned depth and ground truth.
Our model better recovers the fine-grained details and true shape of the objects---this is especially clear in the surface normals extracted from the predictions (last 2 rows of Fig.~\ref{fig:depth-oasis}). Full details, code, and more qualitative results are available \href{https://omnidata.vision}{on our website}.

\noindent\textbf{Surface normal estimation:}
Similar to the existing models on the surface normal track of OASIS, we train a vanilla UNet~\cite{Ronneberger15Unet} architecture (6 down/6 up, similar to \cite{zamir2020consistency}) with angular \& L1 losses, light 2D data augmentation, and input resolutions between 256 and 512. We use Adam~\cite{kingma2014adam} with LR $10^{-4}$ \& weight decay $2\times10^{-6}$. The results in Tab.~\ref{table:oasis_results} indicate that our model matched human-level performance on OASIS $AUC_{p}$. On most of the remaining metrics, it outperformed related models trained on other datasets (including OASIS itself) and models with architectures specifically designed for normals estimation (PBRS). 
Fig.~\ref{fig:surface_normals} shows that our model qualitatively performs \emph{much} better on selected images than is indicated by the numbers, which may be because the standard metrics do not align with perceptual quality as ``uninteresting'' areas (walls, floors) dominate the score~\cite{chen2020oasis}. Further details and results are available in the \href{http://omnidata.vision/supplementary_material/}{supplementary}. 

\begin{figure}[H]
	\vspace{-1mm}
	\centering
	\includegraphics[width=1\columnwidth]{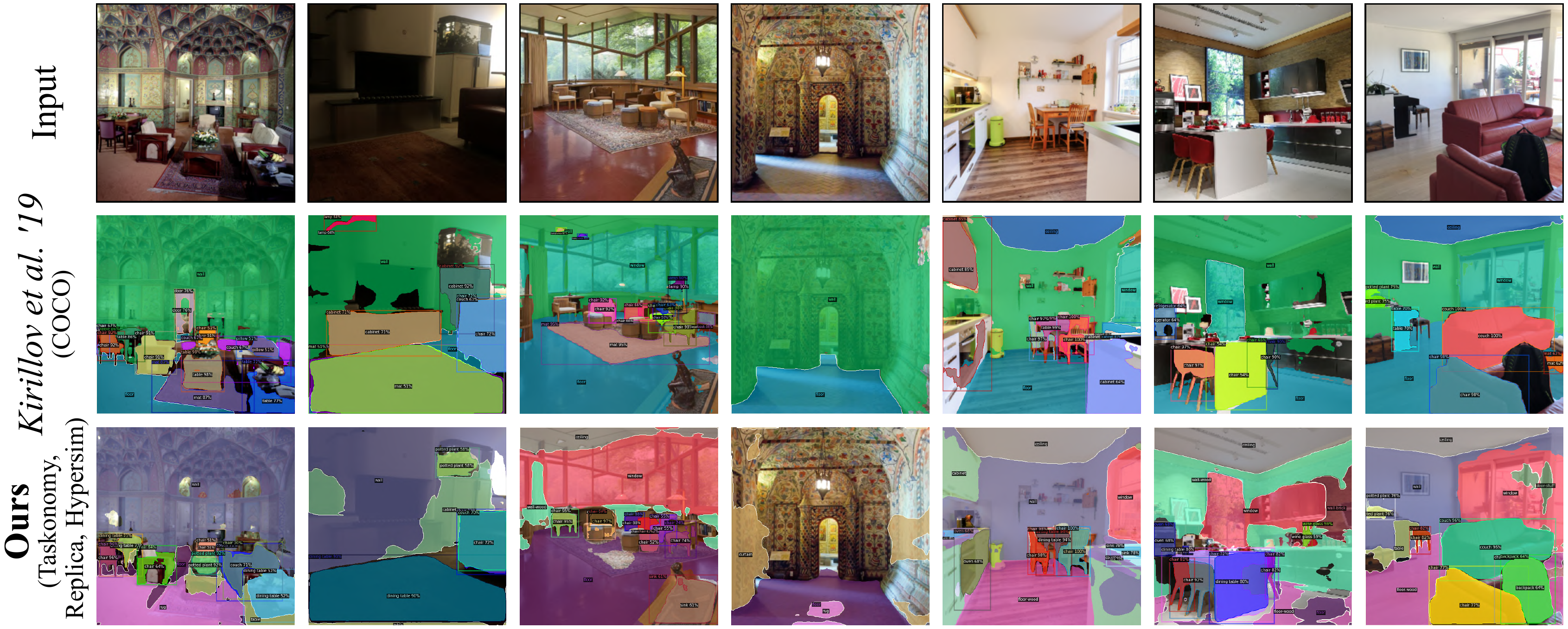}
	\vspace{-6mm}
	\caption{\footnotesize{\textbf{Qualitative results of panoptic segmentation with PanopticFPNs~\cite{Kirillov2019PanopticFP} trained on COCO~\cite{MSCoco} and \handle.}
	The \handle model trained jointly on Taskonomy, Replica, and Hypersim shows good out-of-distribution performance on indoor scenes without people.}}
	\vspace{-6mm}
	\label{fig:panoptic_ood}
\end{figure}

\noindent\textbf{Panoptic segmentation:}
To demonstrate the pipeline's ability to train models for non-geometric tasks, we train a PanopticFPN~\cite{Kirillov2019PanopticFP} on a 3-dataset subset of our starter dataset. Fig.~\ref{fig:panoptic_ood} shows that on in-the-wild images of indoor buildings, the resulting model is of similar quality to one trained on COCO~\cite{MSCoco} (an extensive manually labeled dataset). Quantitative results, full experimental details, and code are available \href{https://omnidata.vision}{on our website}.

\vspace{-0.2cm}
\subsubsection{Dataset Relatedness} \label{sec:internal}
\vspace{-0.2cm}
To estimate how the components of the starter dataset are related, we use zero-shot cross-dataset transfer performance for surface normal and panoptic segmentation models trained on different components. Tab.~\ref{table:inter-dataset} shows that each single model performs well on its corresponding test set, but typically generalizes poorly. The models trained on larger splits perform better overall (see \href{http://omnidata.vision/supplementary_material/}{supplementary}). The model trained on the largest set achieved the best average performance (harmonic mean 25.8\% and 30.3\% better than best single-dataset models for surface normal estimation and panoptic segmentation). The ranking of transfers depended on the task, which might be due to the sparse panoptic labels on Taskonomy (from the followup paper~\cite{armeni_iccv19}), but we believe the dependency is true in general.



\setlength{\tabcolsep}{1pt}
\renewcommand{\arraystretch}{1.3}
\begin{table}[ht]
		\centering
		\footnotesize
		\resizebox{\linewidth}{!}{
		\begin{tabular}{lllllllllll}
			\hline
			
			\multicolumn{1}{|c|}{} &
			\multicolumn{1}{c}{} &
			\multicolumn{4}{c}{Surface normal estimation: L1 Error ($\downarrow$)} &
			\multicolumn{1}{c|}{} &
			\multicolumn{4}{c|}{Panoptic Quality (PQ) ($\uparrow$)}
			\\
			\hline

			\multicolumn{1}{|c|}{Train/Test} &
			\multicolumn{1}{c}{Taskonomy} &
			\multicolumn{1}{c}{Replica} &
			\multicolumn{1}{c}{Hypersim} &
			\multicolumn{1}{c}{Replica+GSO} &
			\multicolumn{1}{c}{BlendedMVG} &
			\multicolumn{1}{c|}{h. mean} &
			\multicolumn{1}{c}{Taskonomy*} &
			\multicolumn{1}{c}{Replica} &
			\multicolumn{1}{c}{Hypersim} &
			\multicolumn{1}{c|}{h. mean}\\
			\hline
			\hline
			\multicolumn{1}{|c|}{Taskonomy*} &
			\multicolumn{1}{c}{\textbf{4.85}} &
			\multicolumn{1}{c}{7.76} &
			\multicolumn{1}{c}{8.69} &
			\multicolumn{1}{c}{13.89} &
			\multicolumn{1}{c}{15.55} &
			\multicolumn{1}{c|}{8.53} &
			\multicolumn{1}{c}{8.39} &
			\multicolumn{1}{c}{3.95} &
			\multicolumn{1}{c}{11.67} &
			\multicolumn{1}{c|}{6.55} \\
			
			\multicolumn{1}{|c|}{Replica} &
			\multicolumn{1}{c}{9.36} &
			\multicolumn{1}{c}{\textbf{3.98}} &
			\multicolumn{1}{c}{11.78} &
			\multicolumn{1}{c}{10.28} &
			\multicolumn{1}{c}{15.02} &
			\multicolumn{1}{c|}{8.24} &
			\multicolumn{1}{c}{1.01} &
			\multicolumn{1}{c}{\textbf{41.97}} &
			\multicolumn{1}{c}{4.50} &
			\multicolumn{1}{c|}{2.43} \\
			\multicolumn{1}{|c|}{Hypersim} &
			\multicolumn{1}{c}{7.28} &
			\multicolumn{1}{c}{7.57} &
			\multicolumn{1}{c}{\textbf{6.72}} &
			\multicolumn{1}{c}{11.34} &
			\multicolumn{1}{c}{12.94} &
			\multicolumn{1}{c|}{8.56} &
			\multicolumn{1}{c}{\textbf{9.35}} &
			\multicolumn{1}{c}{14.08} &
			\multicolumn{1}{c}{25.39} &
			\multicolumn{1}{c|}{13.80} \\
			\multicolumn{1}{|c|}{Replica+GSO} &
			\multicolumn{1}{c}{13.88} &
			\multicolumn{1}{c}{4.94} &
			\multicolumn{1}{c}{15.05} &
			\multicolumn{1}{c}{\textbf{5.17}} &
			\multicolumn{1}{c}{14.03} &
			\multicolumn{1}{c|}{8.26} &
			\multicolumn{1}{c}{-} &
			\multicolumn{1}{c}{-} &
			\multicolumn{1}{c}{-} &
			\multicolumn{1}{c|}{-} \\
			
			\multicolumn{1}{|c|}{BlendedMVG} &
			\multicolumn{1}{c}{17.1} &
			\multicolumn{1}{c}{14.23} &
			\multicolumn{1}{c}{16.93} &
			\multicolumn{1}{c}{14.87} &
			\multicolumn{1}{c}{\textbf{8.85}} &
			\multicolumn{1}{c|}{13.58} &
			\multicolumn{1}{c}{-} &
			\multicolumn{1}{c}{-} &
			\multicolumn{1}{c}{-} &
			\multicolumn{1}{c|}{-} \\
			
			\hline
			\multicolumn{1}{|c|}{\textbf{\handle}} &
			\multicolumn{1}{c}{5.32} &
			\multicolumn{1}{c}{4.24} &
			\multicolumn{1}{c}{6.53} &
			\multicolumn{1}{c}{6.45} &
			\multicolumn{1}{c}{11.53} &
			\multicolumn{1}{c|}{\textbf{6.11}} &
			\multicolumn{1}{c}{9.14} &
			\multicolumn{1}{c}{41.24} &
			\multicolumn{1}{c}{\textbf{30.16}} &
			\multicolumn{1}{c}{\textbf{17.98}} \\

			\hline
		\end{tabular}
		}
        \vspace{-2mm}
		\captionof{table}{\footnotesize{\textbf{Inter-dataset domain transfer performance for surface normal estimation and panoptic segmentation.}
		     Models trained on each individual dataset and \handle are evaluated on test splits of the starter set. The harmonic mean across datasets is shown in the last column.
		     (* PQ on \emph{things} classes only, as Taskonomy does not feature \emph{stuff} labels.)
		}}\label{table:inter-dataset}
    \vspace{-5mm}
\end{table}

\setlength{\tabcolsep}{6pt}
\renewcommand{\arraystretch}{1}

\section{Illustrative Data-Focused Analyses}
\label{sec:baselines}
\vspace{-0.2cm}
Now that we have established that the annotator produces datasets capable of training reliable models, what analyses can we do with such datasets? We survey a few examples here, but they are not intended to be comprehensive (Sec.~\ref{sec:intro}).

\subsection{New 3D Data Augmentations}\label{sec:refocus}
\begin{figure}[H]
    \vspace{-1mm}
    \centering
	\includegraphics[width=0.9\linewidth]{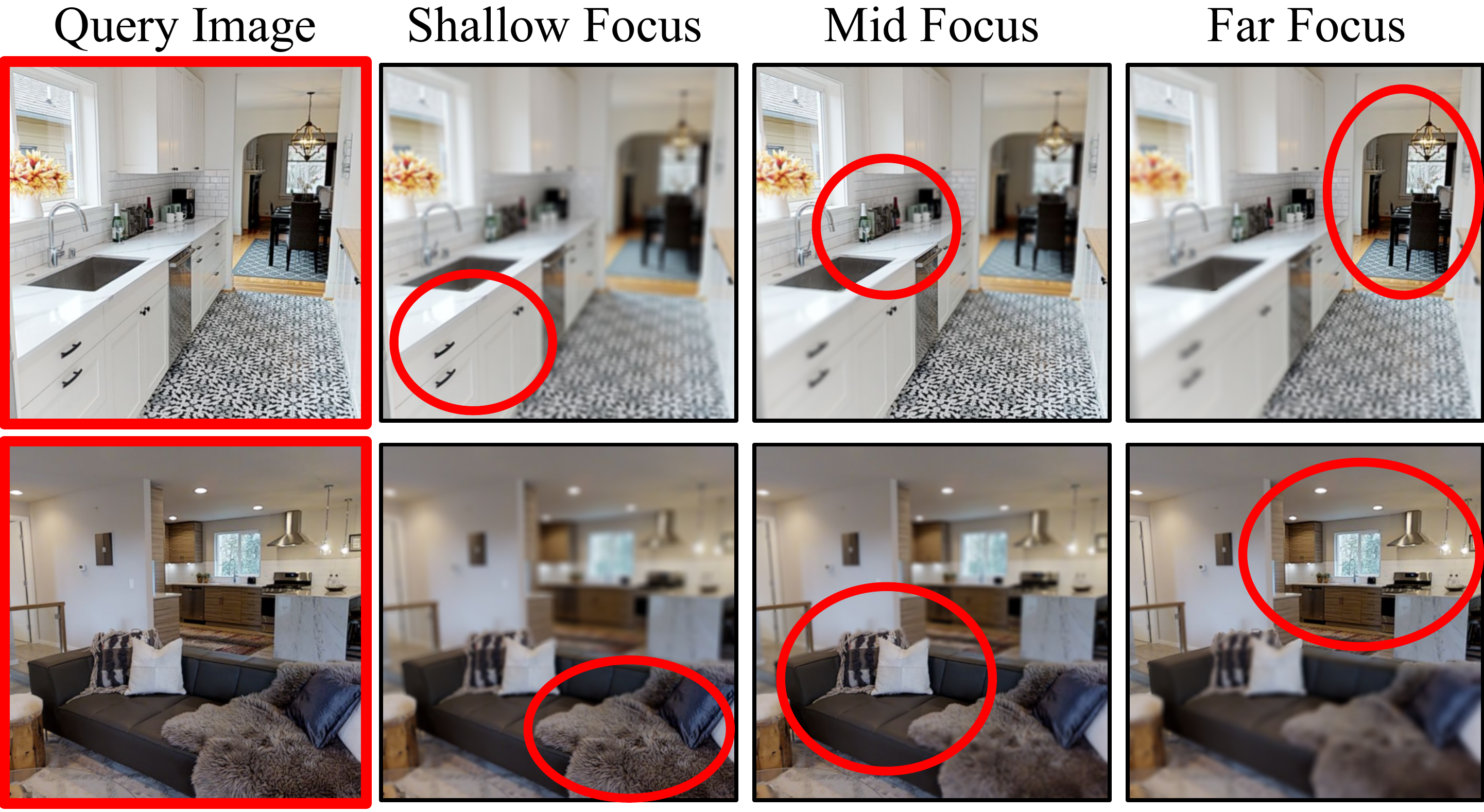}
	\vspace{-2mm}
	\caption{\footnotesize{{\textbf{Image refocusing augmentation on Taskonomy.} Portions of the image that are in focus are highlighted in \textcolor{red}{red} [best viewed zoomed in].}\vspace{-2mm}}}
	\label{fig:blur}
\end{figure}

\vspace{-0.2cm}
Data augmentation is used to address shortcomings in model performance and robustness. For example, models trained only on images captured with narrow apertures (e.g. NYU or Taskonomy) tend to perform poorly on images taken with a wide aperture (i.e. strong depth-of-field), and augmenting with 2D Gaussian blur is used to improve model performance on unfocused portions of images. The approach is common enough that 2D blur was included in the Common Corruptions benchmark~\cite{hendrycks2019benchmarking}. Because the full scene geometry is available for our starter dataset, it is possible to do the \emph{data augmentation in 3D} (image refocusing) instead of 2D (flat blurring). Fig.~\ref{fig:blur} shows an example of what 3D ``image refocusing'' augmentation on our dataset looks like. In the \href{http://omnidata.vision/supplementary_material/}{supplementary}, we show that models trained for surface normal estimation using only 3D augmentation were more robust to both 2D blurring and 3D refocusing than those trained with 2D augmentation.

\subsection{Mid-Level Cues as Inputs: Are They Useful? }\label{sec:mid_level_inputs}
Is there an advantage in using multiple sensors or non-RGB representations of the environment? Instead of predicting mid-level cues as the downstream task (i.e. multi-task learning), multiple cues could also be used as inputs (if relevant sensors are available) or specified as intermediate representation (with the labels being used as supervision only during training, i.e. PADNet~\cite{xu2018pad}). 

Tab.~\ref{replica_semseg_multitask} demonstrates that using these additional cues in the latter 2 ways can improve performance on the original test set and also on unseen data. In this experiment, we train HRNet-18~\cite{sun2019deep} backbones for semantic segmentation using a single component dataset (10 spaces from Replica) and evaluate them on Replica, Hypersim and Taskonomy (tiny split). Relative to using only RGB inputs and the semantic segmentation labels, cross-entropy performance improves across the board when treating the cues as sensors (23\%, 34\% and 30\%) or using them as intermediate representations (13\%, 17\%, and 19\%). Adding more cues seems to help. Full experimental settings in the \href{http://omnidata.vision/supplementary_material/}{supplementary}.

Future work could further analyze how the effectiveness of these different methods change with dataset size, which cues to use, how many additional images a mid-level cue is worth, and on the relative importance of getting more data from the same scene vs. adding data from new scenes.

\setlength{\tabcolsep}{10pt}
\renewcommand{\arraystretch}{1}
\begin{table}[H]
    \addtolength{\tabcolsep}{-4pt}  
    \centering
    \begin{minipage}[]{0.49\textwidth}
		\centering
		\resizebox{\textwidth}{!}{
		\vspace{-5mm}
		\begin{tabular}{|l|llllll|cccccc|}
			\hline
			\multicolumn{1}{|c|}{} & \multicolumn{3}{c|}{GT Mid-Level} & \multicolumn{3}{c|}{Predicted Mid-Level}   \\\cline{2-7}
			\multicolumn{1}{|c|}{Input/Supervised Domains} & \multicolumn{3}{c|}{Cross-Entropy ($\downarrow$)} &  \multicolumn{3}{c|}{Cross-Entropy ($\downarrow$)} \\ 
			\multicolumn{1}{|c|}{} & 
			\multicolumn{1}{c}{Repl.} &
			\multicolumn{1}{c}{H.Sim} &
			\multicolumn{1}{c|}{Task.} &
			\multicolumn{1}{c}{Repl.} &
			\multicolumn{1}{c}{H.Sim} &
			\multicolumn{1}{c|}{Task.} \\ 
			
			\hline
			\hline
			
			\multicolumn{1}{|c|}{RGB} & \multicolumn{1}{c}{0.61} &
			\multicolumn{1}{c}{5.87} &
			\multicolumn{1}{c|}{7.55} &
			
            \multicolumn{1}{c}{0.61} &
			\multicolumn{1}{c}{5.87} &
			\multicolumn{1}{c|}{7.55}
			\\

			\multicolumn{1}{|c|}{(All Above) + Normals} & \multicolumn{1}{c}{0.47} &
			\multicolumn{1}{c}{4.47} &
			\multicolumn{1}{c|}{6.12} &
			\multicolumn{1}{c}{0.61} &
			\multicolumn{1}{c}{5.44} &
			\multicolumn{1}{c|}{7.12}
			\\
			
			\multicolumn{1}{|c|}{(All Above) + 3D Edges} & \multicolumn{1}{c}{\textbf{0.46}} &
			\multicolumn{1}{c}{4.47} &
			\multicolumn{1}{c|}{6.75} &
			\multicolumn{1}{c}{0.54} &
			\multicolumn{1}{c}{5.06} &
			\multicolumn{1}{c|}{6.49}
			\\
			
			\multicolumn{1}{|c|}{(All Above) + (2D Edges, Z-Depth, 3D Keypts)} & \multicolumn{1}{c}{\textbf{0.46}} &
			\multicolumn{1}{c}{\textbf{3.86}} &
			\multicolumn{1}{c|}{\textbf{6.04}} &
			\multicolumn{1}{c}{\textbf{0.53}} &
			\multicolumn{1}{c}{\textbf{4.9}} &
			\multicolumn{1}{c|}{\textbf{6.13}}
            \\
			
			\hline
		\end{tabular}
        		}
        \vspace{-2mm}
		\captionof{table}{\footnotesize{\textbf{Utility of mid-level cues.} The table shows semantic segmentation results using models trained on Replica.
		The models (except for ``RGB") received (either predicted or GT) mid-level cues in their input in addition to the RGB. The results show they notably benefited from the mid-level cues.
		}}\label{replica_semseg_multitask}
	\end{minipage}
    \addtolength{\tabcolsep}{4pt}    
\end{table}
\setlength{\tabcolsep}{6pt}
\renewcommand{\arraystretch}{1}

\subsection{Systematic Evaluation of Multi-Task Learning}\label{sec:mtl_exp}

Recent work~\cite{vandenhende2019branched} shows that existing MTL techniques for computer vision appear to be specialized to their development setting, and in general they do not outperform single-task or shared-encoder approaches on novel datasets or tasks. We extend those results for additional tasks (3D Keypoints) and add numbers on our dataset as a comparison point. Specifically, we follow~\cite{vandenhende2019branched} and train models for a fixed set of tasks (semantic segmentation, 3D keypoints, depth z-buffer, and occlusion edges) using different MTL methods (Tab.~\ref{rank_reversal}). On a 3-dataset split of the starter dataset, some methods naturally perform better and others do worse. One might hope that the ordering of these methods would be the same on different tasks (semantic segmentation vs 3D Keypoints), or at least when training for those same tasks on a different dataset (NYU~\cite{silberman2012indoor}, CityScapes~\cite{cordts2016cityscapes}, or Taskonomy~\cite{taskonomy2018}). Yet, Tab.~\ref{rank_reversal} shows that MTL methods display no clear ranking in either case (i.e. Spearman's $\rho$ was always indistinguishable from 0). Ignoring the lack of significance, the cross-dataset correlation was still weak ($\rho<0.45$), and methods performance was actually \emph{anti-correlated} across tasks ($\rho$=-0.4), suggesting that the models are indeed specialized to specific tasks. This anti-correlation was true even when controlling for dataset.

Given that current MTL approaches do not outperform single-task baselines, predicting different mid-level cues poses a challenging setting for MTL. The \handle pipeline provides an avenue to create large and diverse multi-task mid-level benchmarks that could more systematically and reliably evaluate progress in multi-task learning.

\vspace{-1.0mm}
\renewcommand{\arraystretch}{1.2}
\begin{table}[H]
		\centering
		\addtolength{\tabcolsep}{-4pt}    
		\resizebox{\linewidth}{!}{
		\begin{tabular}{|l|llll|cccccccc|}
			\hline
			\multicolumn{1}{|c|}{} & \multicolumn{8}{c|}{Semantic Segmentation} & \multicolumn{4}{c|}{3D Keypoints}   \\\cline{2-13}
			\multicolumn{1}{|l|}{Method} & \multicolumn{2}{c}{Ours} & \multicolumn{2}{c}{NYU~\cite{silberman2012indoor}} & 
			\multicolumn{2}{c}{CityScapes.~\cite{Vandenhende2020branchedMTL}} &
			\multicolumn{2}{c|}{Taskonomy~\cite{Vandenhende2020branchedMTL}} &
			\multicolumn{2}{c}{Ours} & \multicolumn{2}{c|}{Taskonomy~\cite{Vandenhende2020branchedMTL}} \\ 
			\multicolumn{1}{|c|}{} & 
			\multicolumn{1}{c}{IoU ($\uparrow$)} &
			\multicolumn{1}{c|}{Rank} &
			\multicolumn{1}{c}{IoU ($\uparrow$)} &
			\multicolumn{1}{c|}{Rank} &
			\multicolumn{1}{c}{IoU ($\uparrow$)} &
			\multicolumn{1}{c|}{Rank} &
			\multicolumn{1}{c}{IoU ($\uparrow$)} &
			\multicolumn{1}{c|}{Rank} &
			\multicolumn{1}{c}{L1 ($\downarrow$)} &
			\multicolumn{1}{c|}{Rank} &
			\multicolumn{1}{c}{L1 ($\downarrow$)} &
			\multicolumn{1}{c|}{Rank} \\ 
			
			\hline
			\hline
			
			\multicolumn{1}{|l|}{\hspace{1mm}Single task} & 
			\multicolumn{1}{c}{85.12} &
			\multicolumn{1}{c|}{1} &
			\multicolumn{1}{c}{90.69} &
			\multicolumn{1}{c|}{2}  & 
			\multicolumn{1}{c}{65.2} &
			\multicolumn{1}{c|}{1}  & 
			\multicolumn{1}{c}{43.5}  & 
			\multicolumn{1}{c|}{4} &
			\multicolumn{1}{c}{0.0439} &
			\multicolumn{1}{c|}{4} &
			\multicolumn{1}{c}{0.23} &
			\multicolumn{1}{c|}{1} \\

			\multicolumn{1}{|l|}{\hspace{1mm}MTL baseline} & \multicolumn{1}{c}{81.82} &
			\multicolumn{1}{c|}{3} &
			\multicolumn{1}{c}{90.63} &
			\multicolumn{1}{c|}{3}  & 
			\multicolumn{1}{c}{61.5} &
			\multicolumn{1}{c|}{4}  & 
			\multicolumn{1}{c}{47.8}  & 
			\multicolumn{1}{c|}{1} &
			\multicolumn{1}{c}{0.0429} &
			\multicolumn{1}{c|}{3} &
			\multicolumn{1}{c}{0.34} &
			\multicolumn{1}{c|}{2} \\
			
			\multicolumn{1}{|l|}{\hspace{1mm}MTAN~\cite{liu2019end}} & \multicolumn{1}{c}{83.00} &
			\multicolumn{1}{c|}{2} &
			\multicolumn{1}{c}{91.11} &
			\multicolumn{1}{c|}{1}  & 
			\multicolumn{1}{c}{62.8} &
			\multicolumn{1}{c|}{3}  & 
			\multicolumn{1}{c}{43.8}  & 
			\multicolumn{1}{c|}{3} &
			\multicolumn{1}{c}{0.0426} &
			\multicolumn{1}{c|}{1} &
			\multicolumn{1}{c}{0.4} &
			\multicolumn{1}{c|}{3} \\
			
			\multicolumn{1}{|l|}{\hspace{1mm}Cross-stitch~\cite{misra2016cross}} & \multicolumn{1}{c}{80.69} &
			\multicolumn{1}{c|}{4} &
			\multicolumn{1}{c}{90.33} &
			\multicolumn{1}{c|}{4}  & 
			\multicolumn{1}{c}{65.1} &
			\multicolumn{1}{c|}{2}  & 
			\multicolumn{1}{c}{44.0}  & 
			\multicolumn{1}{c|}{2} &
			\multicolumn{1}{c}{0.0427} &
			\multicolumn{1}{c|}{2} &
			\multicolumn{1}{c}{0.50} &
			\multicolumn{1}{c|}{4} \\
			
			\hline
				
			\multicolumn{1}{|l|}{Spearman's $\rho$} & \multicolumn{8}{c|}{
			Within task.: $\rho{=}$0.43.  Between segm.-3D keypts.: $\rho{=}$-0.4
			}  & 
			\multicolumn{4}{c|}{
			Within task: $\rho{=}$0.2.
			}  \\
			
			\hline
		\end{tabular}
		}
		\vspace{-2mm}
		\captionof{table}{\footnotesize{\textbf{Multi-task training methods do not show a clear ordering.} Within-task, rankings between different methods were indistinguishable from random orderings (i.e. $\rho{=}$0). Between tasks, rankings on Sem. Seg. were anti-correlated with rank on the 3D Keypts ($\rho{=}$-0.4). Both conclusions were strengthened after controlling for training setups.}
		}\label{rank_reversal}
	\addtolength{\tabcolsep}{4pt}    
\vspace{-3mm}
\end{table}
\renewcommand{\arraystretch}{1}

\section{Conclusion and Limitations}
This paper introduces a pipeline to create steerable datasets from comprehensive scans of the environment. The resulting multi-task datasets can be large and diverse, and realistic enough that models trained on the data perform well in the real world. To demonstrate this, we annotated an example dataset and used it to train a few standard vision methods to state-of-the-art performance on multiple computer vision tasks. We believe this is capability is useful on its own, especially since it acts as a bridge between real-world 3D scans, simulators, and static vision datasets. 

Our main intention for this tool is to better study properties of vision datasets, and their interaction with models and tasks. Crucially, the fact that the pipeline can be used to train strong models in real-world settings gives us hope that findings stemming from this pipeline here might hold true more generally. In particular, we believe that this ``steerable dataset'' method could bear fruit in fundamental lines of research such as how data sampling strategies and choice of cue/sensor impact representations and model reliability. 

To close, we discuss some of important limitations of this pipeline and possibilities for future lines of work.

\begin{enumerate}[leftmargin=4mm]
    \item \textbf{Studying steerability.} The Omnidata pipeline provides a method to create steerable datasets, but we did not present any analysis of the effects of tuning the different steering `knobs'. I.e. our starter dataset used a fixed choice of generation settings and we did no tuning on that initial choice.  Clearly, such choices do have important effects on the dataset (e.g. see our online \href{https://omnidata.vision/designer}{demo}) and on the resulting models~\cite{Vandenhende2020branchedMTL, unbiasedLook2011Torralba, bias2019Azulay, pascal2014xiang}). We believe rich insights lie in this direction, which is why we created this pipeline. This paper only provides a few sporadic experiments to illustrate the general idea, it does not represent a systematic study. 
    
    \item \textbf{Limited capture information.} The 3D scans used in this paper come from the output of standard structure-from-motion methods that stitch together many overlapping images from RGB and depth sensors. These scans are represented as meshes (with textures or aligned RGB images), but this representation leaves out important information about the scene. For example, the materials lack reflectance models (e.g. BRDF) and there is no information about scene lighting. Moreover, scans usually have limited reconstruction accuracy (e.g. commonly up to 2cm error in Taskonomy), which affects both the texture quality and the quality of the generated labels. Better sensing technology (e.g. light-field cameras, higher-resolution depth sensors), as well as algorithmic improvements (e.g. NeRF, below) can add more dimensions of control and reduce the gap between the resampled and real cues/images.
    
    \item \textbf{How to represent the 'complete capture'.} The Omnidata pipeline uses 3D meshes to represent the scene, and samples images using that representation. Other representations, such as using light-field cameras and NeRF~\cite{mildenhall2020nerf} could be used as implicit representations of the scene and similarly used for resampling the scene. The surprising effectiveness of NeRF makes this direction quite compelling.
    
    \item \textbf{Limited number of mid-level cues.} The initial release of the Omnidata annotator provides 21 mid-level cues. Like most tasks in computer vision, the current mid-level cues are based more on human intuition than on demonstrably predictive theories of vision. As computer vision and vision science make new advancements, these can be integrated to the sampling pipeline as long as the required information is present in the capture information (e.g. new cues and augmentations).
\end{enumerate}

{\small
\bibliographystyle{ieee_fullname}
\bibliography{egbib}

\begin{thebibliography}{10}\itemsep=-1pt

\bibitem{gsoignition}
Ignition app.

\bibitem{armeni_iccv19}
Iro Armeni, Zhi-Yang He, JunYoung Gwak, Amir~R. Zamir, Martin Fischer, Jitendra
  Malik, and Silvio Savarese.
\newblock 3d scene graph: A structure for unified semantics, 3d space, and
  camera.
\newblock In {\em Proceedings of the IEEE International Conference on Computer
  Vision}, 2019.

\bibitem{2d3ds}
Iro Armeni, Sasha Sax, Amir~Roshan Zamir, and Silvio Savarese.
\newblock Joint 2d-3d-semantic data for indoor scene understanding.
\newblock {\em CoRR}, abs/1702.01105, 2017.

\bibitem{armeni2017joint}
Iro Armeni, Sasha Sax, Amir~R Zamir, and Silvio Savarese.
\newblock Joint 2d-3d-semantic data for indoor scene understanding.
\newblock {\em arXiv preprint arXiv:1702.01105}, 2017.

\bibitem{armeni20163d}
Iro Armeni, Ozan Sener, Amir~R Zamir, Helen Jiang, Ioannis Brilakis, Martin
  Fischer, and Silvio Savarese.
\newblock 3d semantic parsing of large-scale indoor spaces.
\newblock In {\em Proceedings of the IEEE Conference on Computer Vision and
  Pattern Recognition}, pages 1534--1543, 2016.

\bibitem{azulay2018deep}
Aharon Azulay and Yair Weiss.
\newblock Why do deep convolutional networks generalize so poorly to small
  image transformations?
\newblock {\em arXiv preprint arXiv:1805.12177}, 2018.

\bibitem{bias2019Azulay}
Aharon Azulay and Yair Weiss.
\newblock Why do deep convolutional networks generalize so poorly to small
  image transformations?
\newblock {\em CoRR}, abs/1805.12177, 2018.

\bibitem{barbu2019objectnet}
Andrei Barbu, David Mayo, Julian Alverio, William Luo, Christopher Wang, Danny
  Gutfreund, Joshua Tenenbaum, and Boris Katz.
\newblock Objectnet: A large-scale bias-controlled dataset for pushing the
  limits of object recognition models.
\newblock 2019.

\bibitem{calli2015benchmarking}
Berk Calli, Aaron Walsman, Arjun Singh, Siddhartha Srinivasa, Pieter Abbeel,
  and Aaron~M Dollar.
\newblock Benchmarking in manipulation research: The ycb object and model set
  and benchmarking protocols.
\newblock {\em arXiv preprint arXiv:1502.03143}, 2015.

\bibitem{Matterport3D}
Angel Chang, Angela Dai, Thomas Funkhouser, Maciej Halber, Matthias Niessner,
  Manolis Savva, Shuran Song, Andy Zeng, and Yinda Zhang.
\newblock Matterport3d: Learning from rgb-d data in indoor environments.
\newblock {\em International Conference on 3D Vision (3DV)}, 2017.

\bibitem{chen2020simple}
Ting Chen, Simon Kornblith, Mohammad Norouzi, and Geoffrey Hinton.
\newblock A simple framework for contrastive learning of visual
  representations, 2020.

\bibitem{Chen16MonocularDepth}
Weifeng Chen, Zhao Fu, Dawei Yang, and Jia Deng.
\newblock Single-image depth perception in the wild.
\newblock {\em CoRR}, abs/1604.03901, 2016.

\bibitem{chen2020oasis}
Weifeng Chen, Shengyi Qian, David Fan, Noriyuki Kojima, Max Hamilton, and Jia
  Deng.
\newblock Oasis: A large-scale dataset for single image 3d in the wild, 2020.

\bibitem{Chen17Snow}
Weifeng Chen, Donglai Xiang, and Jia Deng.
\newblock Surface normals in the wild.
\newblock {\em CoRR}, abs/1704.02956, 2017.

\bibitem{cignoni2008meshlab}
Paolo Cignoni, Marco Callieri, Massimiliano Corsini, Matteo Dellepiane, Fabio
  Ganovelli, and Guido Ranzuglia.
\newblock Meshlab: an open-source mesh processing tool.
\newblock In {\em Eurographics Italian chapter conference}, volume 2008, pages
  129--136. Salerno, Italy, 2008.

\bibitem{community2018blender}
BO Community.
\newblock Blender--a 3d modelling and rendering package.
\newblock 2018.

\bibitem{blender3d}
Blender~Online Community.
\newblock {\em Blender - a 3D modelling and rendering package}.
\newblock Blender Foundation, Stichting Blender Foundation, Amsterdam, 2018.

\bibitem{cordts2016cityscapes}
Marius Cordts, Mohamed Omran, Sebastian Ramos, Timo Rehfeld, Markus Enzweiler,
  Rodrigo Benenson, Uwe Franke, Stefan Roth, and Bernt Schiele.
\newblock The cityscapes dataset for semantic urban scene understanding.
\newblock In {\em Proceedings of the IEEE conference on computer vision and
  pattern recognition}, pages 3213--3223, 2016.

\bibitem{imagenet}
J. Deng, W. Dong, R. Socher, L.-J. Li, K. Li, and L. Fei-Fei.
\newblock {ImageNet: A Large-Scale Hierarchical Image Database}.
\newblock In {\em CVPR09}, 2009.

\bibitem{denninger2019blenderproc}
Maximilian Denninger, Martin Sundermeyer, Dominik Winkelbauer, Youssef Zidan,
  Dmitry Olefir, Mohamad Elbadrawy, Ahsan Lodhi, and Harinandan Katam.
\newblock Blenderproc.
\newblock {\em arXiv preprint arXiv:1911.01911}, 2019.

\bibitem{Dosovitskiy17}
Alexey Dosovitskiy, German Ros, Felipe Codevilla, Antonio Lopez, and Vladlen
  Koltun.
\newblock {CARLA}: {An} open urban driving simulator.
\newblock In {\em Proceedings of the 1st Annual Conference on Robot Learning},
  pages 1--16, 2017.

\bibitem{Gibson1966}
J. Gibson.
\newblock The senses considered as perceptual systems.
\newblock 1966.

\bibitem{grill2020bootstrap}
Jean-Bastien Grill, Florian Strub, Florent Altché, Corentin Tallec, Pierre~H.
  Richemond, Elena Buchatskaya, Carl Doersch, Bernardo~Avila Pires,
  Zhaohan~Daniel Guo, Mohammad~Gheshlaghi Azar, Bilal Piot, Koray Kavukcuoglu,
  Rémi Munos, and Michal Valko.
\newblock Bootstrap your own latent: A new approach to self-supervised
  learning, 2020.

\bibitem{hendrycks2019benchmarking}
Dan Hendrycks and Thomas Dietterich.
\newblock Benchmarking neural network robustness to common corruptions and
  perturbations.
\newblock {\em arXiv preprint arXiv:1903.12261}, 2019.

\bibitem{hinton2015distilling}
Geoffrey Hinton, Oriol Vinyals, and Jeff Dean.
\newblock Distilling the knowledge in a neural network.
\newblock {\em arXiv preprint arXiv:1503.02531}, 2015.

\bibitem{mirrorflow}
Junhwa Hur and Stefan Roth.
\newblock Mirrorflow: Exploiting symmetries in joint optical flow and occlusion
  estimation.
\newblock {\em CoRR}, abs/1708.05355, 2017.

\bibitem{leftRightConsistency}
Zequn Jie, Pengfei Wang, Yonggen Ling, Bo Zhao, Yunchao Wei, Jiashi Feng, and
  Wei Liu.
\newblock Left-right comparative recurrent model for stereo matching.
\newblock {\em CoRR}, abs/1804.00796, 2018.

\bibitem{johnson2017clevr}
Justin Johnson, Bharath Hariharan, Laurens Van Der~Maaten, Li Fei-Fei, C
  Lawrence~Zitnick, and Ross Girshick.
\newblock Clevr: A diagnostic dataset for compositional language and elementary
  visual reasoning.
\newblock In {\em Proceedings of the IEEE conference on computer vision and
  pattern recognition}, pages 2901--2910, 2017.

\bibitem{vizdoom}
Michal Kempka, Marek Wydmuch, Grzegorz Runc, Jakub Toczek, and Wojciech
  Jaskowski.
\newblock Vizdoom: {A} doom-based {AI} research platform for visual
  reinforcement learning.
\newblock {\em arXiv preprint arXiv:1605.02097}, 2016.

\bibitem{kingma2014adam}
Diederik~P. Kingma and Jimmy Ba.
\newblock Adam: {A} method for stochastic optimization.
\newblock {\em CoRR}, abs/1412.6980, 2014.

\bibitem{Kirillov2019PanopticFP}
Alexander Kirillov, Ross~B. Girshick, Kaiming He, and Piotr Doll{\'a}r.
\newblock Panoptic feature pyramid networks.
\newblock {\em 2019 IEEE/CVF Conference on Computer Vision and Pattern
  Recognition (CVPR)}, pages 6392--6401, 2019.

\bibitem{Knapitsch2017}
Arno Knapitsch, Jaesik Park, Qian-Yi Zhou, and Vladlen Koltun.
\newblock Tanks and temples: Benchmarking large-scale scene reconstruction.
\newblock {\em ACM Transactions on Graphics}, 36(4), 2017.

\bibitem{knapitsch2017tanks}
Arno Knapitsch, Jaesik Park, Qian-Yi Zhou, and Vladlen Koltun.
\newblock Tanks and temples: Benchmarking large-scale scene reconstruction.
\newblock {\em ACM Transactions on Graphics (ToG)}, 36(4):1--13, 2017.

\bibitem{lake2019omniglot}
Brenden~M Lake, Ruslan Salakhutdinov, and Joshua~B Tenenbaum.
\newblock The omniglot challenge: a 3-year progress report.
\newblock {\em Current Opinion in Behavioral Sciences}, 29:97--104, 2019.

\bibitem{lambert2020mseg}
John Lambert, Zhuang Liu, Ozan Sener, James Hays, and Vladlen Koltun.
\newblock Mseg: a composite dataset for multi-domain semantic segmentation.
\newblock In {\em Proceedings of the IEEE/CVF Conference on Computer Vision and
  Pattern Recognition}, pages 2879--2888, 2020.

\bibitem{MSCoco}
Tsung{-}Yi Lin, Michael Maire, Serge~J. Belongie, Lubomir~D. Bourdev, Ross~B.
  Girshick, James Hays, Pietro Perona, Deva Ramanan, Piotr Doll{\'{a}}r, and
  C.~Lawrence Zitnick.
\newblock Microsoft {COCO:} common objects in context.
\newblock {\em CoRR}, abs/1405.0312, 2014.

\bibitem{liu2019end}
Shikun Liu, Edward Johns, and Andrew~J Davison.
\newblock End-to-end multi-task learning with attention.
\newblock In {\em Proceedings of the IEEE Conference on Computer Vision and
  Pattern Recognition}, pages 1871--1880, 2019.

\bibitem{liu2015faceattributes}
Ziwei Liu, Ping Luo, Xiaogang Wang, and Xiaoou Tang.
\newblock Deep learning face attributes in the wild.
\newblock In {\em Proceedings of International Conference on Computer Vision
  (ICCV)}, December 2015.

\bibitem{maltoni2018icifar}
Davide Maltoni and Vincenzo Lomonaco.
\newblock Continuous learning in single-incremental-task scenarios.
\newblock {\em CoRR}, abs/1806.08568, 2018.

\bibitem{habitat19iccv}
{Manolis Savva*}, {Abhishek Kadian*}, {Oleksandr Maksymets*}, Yili Zhao, Erik
  Wijmans, Bhavana Jain, Julian Straub, Jia Liu, Vladlen Koltun, Jitendra
  Malik, Devi Parikh, and Dhruv Batra.
\newblock Habitat: {A} {P}latform for {E}mbodied {AI} {R}esearch.
\newblock In {\em Proceedings of the IEEE/CVF International Conference on
  Computer Vision (ICCV)}, 2019.

\bibitem{mildenhall2020nerf}
Ben Mildenhall, Pratul~P Srinivasan, Matthew Tancik, Jonathan~T Barron, Ravi
  Ramamoorthi, and Ren Ng.
\newblock Nerf: Representing scenes as neural radiance fields for view
  synthesis.
\newblock In {\em European Conference on Computer Vision}, pages 405--421.
  Springer, 2020.

\bibitem{misra2016cross}
Ishan Misra, Abhinav Shrivastava, Abhinav Gupta, and Martial Hebert.
\newblock Cross-stitch networks for multi-task learning.
\newblock In {\em Proceedings of the IEEE Conference on Computer Vision and
  Pattern Recognition}, pages 3994--4003, 2016.

\bibitem{mnih-dqn-2015}
Volodymyr Mnih, Koray Kavukcuoglu, David Silver, Andrei~A. Rusu, Joel Veness,
  Marc~G. Bellemare, Alex Graves, Martin Riedmiller, Andreas~K. Fidjeland,
  Georg Ostrovski, Stig Petersen, Charles Beattie, Amir Sadik, Ioannis
  Antonoglou, Helen King, Dharshan Kumaran, Daan Wierstra, Shane Legg, and
  Demis Hassabis.
\newblock Human-level control through deep reinforcement learning.
\newblock {\em Nature}, 518(7540):529--533, 02 2015.

\bibitem{NEURIPS2019_9015}
Adam Paszke, Sam Gross, Francisco Massa, Adam Lerer, James Bradbury, Gregory
  Chanan, Trevor Killeen, Zeming Lin, Natalia Gimelshein, Luca Antiga, Alban
  Desmaison, Andreas Kopf, Edward Yang, Zachary DeVito, Martin Raison, Alykhan
  Tejani, Sasank Chilamkurthy, Benoit Steiner, Lu Fang, Junjie Bai, and Soumith
  Chintala.
\newblock Pytorch: An imperative style, high-performance deep learning library.
\newblock In H. Wallach, H. Larochelle, A. Beygelzimer, F. d\textquotesingle
  Alch\'{e}-Buc, E. Fox, and R. Garnett, editors, {\em Advances in Neural
  Information Processing Systems 32}, pages 8024--8035. Curran Associates,
  Inc., 2019.

\bibitem{Purushwalkam2020Demystifying}
Senthil Purushwalkam Shiva~Prakash and Abhinav Gupta.
\newblock Demystifying contrastive self-supervised learning: Invariances,
  augmentations and dataset biases.
\newblock {\em Advances in Neural Information Processing Systems}, 33, 2020.

\bibitem{Ranftl2021}
Ren\'{e} Ranftl, Alexey Bochkovskiy, and Vladlen Koltun.
\newblock Vision transformers for dense prediction.
\newblock {\em ArXiv preprint}, 2021.

\bibitem{ranftl2019towards}
Ren{\'e} Ranftl, Katrin Lasinger, David Hafner, Konrad Schindler, and Vladlen
  Koltun.
\newblock Towards robust monocular depth estimation: Mixing datasets for
  zero-shot cross-dataset transfer.
\newblock {\em arXiv preprint arXiv:1907.01341}, 2019.

\bibitem{Richter_2016_ECCV}
Stephan~R. Richter, Vibhav Vineet, Stefan Roth, and Vladlen Koltun.
\newblock Playing for data: {G}round truth from computer games.
\newblock In Bastian Leibe, Jiri Matas, Nicu Sebe, and Max Welling, editors,
  {\em European Conference on Computer Vision (ECCV)}, volume 9906 of {\em
  LNCS}, pages 102--118. Springer International Publishing, 2016.

\bibitem{roberts:2020}
Mike Roberts and Nathan Paczan.
\newblock {Hypersim}: {A} photorealistic synthetic dataset for holistic indoor
  scene understanding.
\newblock arXiv 2020.

\bibitem{Ronneberger15Unet}
Olaf Ronneberger, Philipp Fischer, and Thomas Brox.
\newblock U-net: Convolutional networks for biomedical image segmentation.
\newblock {\em CoRR}, abs/1505.04597, 2015.

\bibitem{schoenberger2016sfm}
Johannes~Lutz Sch\"{o}nberger and Jan-Michael Frahm.
\newblock Structure-from-motion revisited.
\newblock In {\em Conference on Computer Vision and Pattern Recognition
  (CVPR)}, 2016.

\bibitem{selftraining}
H. Scudder.
\newblock Probability of error of some adaptive pattern-recognition machines.
\newblock {\em IEEE Transactions on Information Theory}, 11(3):363--371, 1965.

\bibitem{silberman2012indoor}
Nathan Silberman, Derek Hoiem, Pushmeet Kohli, and Rob Fergus.
\newblock Indoor segmentation and support inference from rgbd images.
\newblock In {\em European conference on computer vision}, pages 746--760.
  Springer, 2012.

\bibitem{Silberman2012}
Nathan Silberman, Derek Hoiem, Pushmeet Kohli, and Rob Fergus.
\newblock {\em Indoor Segmentation and Support Inference from RGBD Images},
  pages 746--760.
\newblock Springer Berlin Heidelberg, Berlin, Heidelberg, 2012.

\bibitem{song2016ssc}
Shuran Song, Fisher Yu, Andy Zeng, Angel~X Chang, Manolis Savva, and Thomas
  Funkhouser.
\newblock Semantic scene completion from a single depth image.
\newblock {\em Proceedings of 30th IEEE Conference on Computer Vision and
  Pattern Recognition}, 2017.

\bibitem{replica19arxiv}
Julian Straub, Thomas Whelan, Lingni Ma, Yufan Chen, Erik Wijmans, Simon Green,
  Jakob~J. Engel, Raul Mur-Artal, Carl Ren, Shobhit Verma, Anton Clarkson,
  Mingfei Yan, Brian Budge, Yajie Yan, Xiaqing Pan, June Yon, Yuyang Zou,
  Kimberly Leon, Nigel Carter, Jesus Briales, Tyler Gillingham, Elias Mueggler,
  Luis Pesqueira, Manolis Savva, Dhruv Batra, Hauke~M. Strasdat, Renzo~De
  Nardi, Michael Goesele, Steven Lovegrove, and Richard Newcombe.
\newblock The {R}eplica dataset: A digital replica of indoor spaces.
\newblock {\em arXiv preprint arXiv:1906.05797}, 2019.

\bibitem{Sun2017RevisitingUE}
Chen Sun, Abhinav Shrivastava, Saurabh Singh, and A. Gupta.
\newblock Revisiting unreasonable effectiveness of data in deep learning era.
\newblock {\em 2017 IEEE International Conference on Computer Vision (ICCV)},
  pages 843--852, 2017.

\bibitem{sun2019deep}
Ke Sun, Bin Xiao, Dong Liu, and Jingdong Wang.
\newblock Deep high-resolution representation learning for human pose
  estimation.
\newblock In {\em Proceedings of the IEEE/CVF Conference on Computer Vision and
  Pattern Recognition}, pages 5693--5703, 2019.

\bibitem{tobin2017domain}
Josh Tobin, Rachel Fong, Alex Ray, Jonas Schneider, Wojciech Zaremba, and
  Pieter Abbeel.
\newblock Domain randomization for transferring deep neural networks from
  simulation to the real world.
\newblock In {\em 2017 IEEE/RSJ International Conference on Intelligent Robots
  and Systems (IROS)}, pages 23--30. IEEE, 2017.

\bibitem{unbiasedLook2011Torralba}
Antonio Torralba and Alexei~A. Efros.
\newblock Unbiased look at dataset bias.
\newblock In {\em CVPR 2011}, pages 1521--1528, 2011.

\bibitem{Vandenhende2020branchedMTL}
Simon Vandenhende, Bert~De Brabandere, and Luc~Van Gool.
\newblock Branched multi-task networks: Deciding what layers to share.
\newblock {\em CoRR}, abs/1904.02920, 2019.

\bibitem{vandenhende2019branched}
Simon Vandenhende, Stamatios Georgoulis, Bert De~Brabandere, and Luc Van~Gool.
\newblock Branched multi-task networks: deciding what layers to share.
\newblock {\em arXiv preprint arXiv:1904.02920}, 2019.

\bibitem{vandenhende2021multi}
Simon Vandenhende, Stamatios Georgoulis, Wouter Van~Gansbeke, Marc Proesmans,
  Dengxin Dai, and Luc Van~Gool.
\newblock Multi-task learning for dense prediction tasks: A survey.
\newblock {\em IEEE Transactions on Pattern Analysis and Machine Intelligence},
  2021.

\bibitem{VinyalsBLKW16Matching}
Oriol Vinyals, Charles Blundell, Timothy~P. Lillicrap, Koray Kavukcuoglu, and
  Daan Wierstra.
\newblock Matching networks for one shot learning.
\newblock {\em CoRR}, abs/1606.04080, 2016.

\bibitem{WelinderEtal2010}
P. Welinder, S. Branson, T. Mita, C. Wah, F. Schroff, S. Belongie, and P.
  Perona.
\newblock {Caltech-UCSD Birds 200}.
\newblock Technical Report CNS-TR-2010-001, California Institute of Technology,
  2010.

\bibitem{gibson}
Fei Xia, Amir Zamir, Zhi-Yang He, Alexander Sax, Jitendra Malik, and Silvio
  Savarese.
\newblock {Gibson Env}: Real-world perception for embodied agents.
\newblock In {\em 2018 IEEE Conference on Computer Vision and Pattern
  Recognition (CVPR)}, 2018.

\bibitem{pascal2014xiang}
Yu Xiang, Roozbeh Mottaghi, and Silvio Savarese.
\newblock Beyond pascal: A benchmark for 3d object detection in the wild.
\newblock In {\em IEEE Winter Conference on Applications of Computer Vision},
  pages 75--82, 2014.

\bibitem{xu2018pad}
Dan Xu, Wanli Ouyang, Xiaogang Wang, and Nicu Sebe.
\newblock Pad-net: Multi-tasks guided prediction-and-distillation network for
  simultaneous depth estimation and scene parsing.
\newblock In {\em Proceedings of the IEEE Conference on Computer Vision and
  Pattern Recognition}, pages 675--684, 2018.

\bibitem{yao2020blendedmvs}
Yao Yao, Zixin Luo, Shiwei Li, Jingyang Zhang, Yufan Ren, Lei Zhou, Tian Fang,
  and Long Quan.
\newblock Blendedmvs: A large-scale dataset for generalized multi-view stereo
  networks.
\newblock {\em Computer Vision and Pattern Recognition (CVPR)}, 2020.

\bibitem{yin2018geonet}
Zhichao Yin and Jianping Shi.
\newblock Geonet: Unsupervised learning of dense depth, optical flow and camera
  pose.
\newblock In {\em Proceedings of the IEEE conference on computer vision and
  pattern recognition}, pages 1983--1992, 2018.

\bibitem{zamir2020consistency}
Amir Zamir, Alexander Sax, Teresa Yeo, Oğuzhan Kar, Nikhil Cheerla, Rohan
  Suri, Zhangjie Cao, Jitendra Malik, and Leonidas Guibas.
\newblock Robust learning through cross-task consistency.
\newblock {\em arXiv}, 2020.

\bibitem{taskonomy2018}
Amir~R. Zamir, Alexander Sax, William~B. Shen, Leonidas~J. Guibas, Jitendra
  Malik, and Silvio Savarese.
\newblock Taskonomy: Disentangling task transfer learning.
\newblock In {\em IEEE Conference on Computer Vision and Pattern Recognition
  (CVPR)}. IEEE, 2018.

\bibitem{zhang2019sidetuning}
Jeffrey~O Zhang, Alexander Sax, Amir Zamir, Leonidas Guibas, and Jitendra
  Malik.
\newblock Side-tuning: A baseline for network adaptation via additive side
  networks, 2019.

\bibitem{Zhang16PBRS}
Yinda Zhang, Shuran Song, Ersin Yumer, Manolis Savva, Joon{-}Young Lee, Hailin
  Jin, and Thomas~A. Funkhouser.
\newblock Physically-based rendering for indoor scene understanding using
  convolutional neural networks.
\newblock {\em CoRR}, abs/1612.07429, 2016.

\end{thebibliography}
}

\end{document}